\RequirePackage{luatex85}
\def\year{2019}\relax
\documentclass[letterpaper]{article} 
\usepackage{aaai19}  
\usepackage{times}  
\usepackage{helvet}  
\usepackage{courier}  
\usepackage{url}  
\usepackage{graphicx}  

\usepackage{subcaption}
\usepackage{amsmath}
\usepackage{amssymb}
\usepackage{amsfonts}
\usepackage{threeparttable}
\usepackage{multirow}

\begin{document}
%
\title{Representation Learning by Reconstructing Neighborhoods}
\author{
Chin-Chia Michael Yeh, Yan Zhu, Evangelos E. Papalexakis, $\dagger$ Abdullah Mueen, Eamonn Keogh\\
University of California, Riverside, $\dagger$ University of New Mexico\\
\{myeh003, yzhu015\}@ucr.edu, epapalex@cs.ucr.edu, mueen@unm.edu, eamonn@cs.ucr.edu\\
}
\maketitle
\begin{abstract}
Since its introduction, unsupervised representation learning has attracted a lot of attention from the research community, as it is demonstrated to be highly effective and easy-to-apply in tasks such as dimension reduction, clustering, visualization, information retrieval, and semi-supervised learning. 
In this work, we propose a novel unsupervised representation learning framework called \emph{neighbor-encoder}, in which domain knowledge can be easily incorporated into the learning process without modifying the general encoder-decoder architecture of the classic autoencoder.
In contrast to autoencoder, which reconstructs the input data \emph{itself}, neighbor-encoder reconstructs the input data's \emph{neighbors}. 
As the proposed representation learning problem is essentially a neighbor reconstruction problem, domain knowledge can be easily incorporated in the form of an appropriate definition of similarity between objects. 
Based on that observation, our framework can leverage any off-the-shelf similarity search algorithms or side information to find the neighbor of an input object. 
Applications of other algorithms (e.g., association rule mining) in our framework are also possible, given that the appropriate definition of \emph{neighbor} can vary in different contexts. 
We have demonstrated the effectiveness of our framework in many diverse domains, including images, text, and time series, and for various data mining tasks including classification, clustering, and visualization. 
Experimental results show that neighbor-encoder not only outperforms autoencoder in most of the scenarios we consider, but also achieves the state-of-the-art performance on text document clustering.
\end{abstract}

\section{Introduction}
Unsupervised representation learning has been shown effective in tasks such as dimension reduction, clustering, visualization, information retrieval, and semi-supervised learning~\cite{goodfellow2016book}.
Learned representations have been shown to achieve better performance on individual tasks than domain-speciﬁc handcrafted features, and different tasks can use the same learned representation~\cite{goodfellow2016book}.
For example, the embedding obtained by methods like word2vec~\cite{mikolov2013nips} has been exploited in many different text mining systems~\cite{catherine2017transnets,zheng2017joint}.
Moreover, to help a user extract knowledge from a data set, a data exploration system can first learn the representation without supervision for each item in the data set; then display both the clustering (e.g., $k$-means~\cite{lloyd1982tit}) and visualization (e.g., $2D$ projection with $t$-Distributed Stochastic Neighbor Embedding/$t$-SNE~\cite{maaten2008jmlr}) results produced from the representation.

There are two types of unsupervised representation learning methods: \emph{domain-specific} unsupervised representation learning methods and \emph{general} unsupervised representation learning methods.
While domain-specific unsupervised representation learning methods like word2vec~\cite{mikolov2013nips} and video-based methods~\cite{agrawal2015iccv,jayaraman2015iccv,wang2015iccv,pathak2017cvpr} have been widely adopted in their respective domains, their success cannot be directly transferred to other domains because their assumptions do not hold for other types of data.
In contrast, general unsupervised representation learning methods, such as autoencoder~\cite{bengio2007nips,huang2007cvpr,vincent2010jmlr}, can be effortlessly applied to data from various domains, but the performance of general methods is usually inferior to those that utilize domain knowledge~\cite{mikolov2013nips,agrawal2015iccv,jayaraman2015iccv,wang2015iccv,pathak2017cvpr}.

\begin{figure}[htb]
\centering
\includegraphics[trim={9.9cm 0cm 10cm 0cm}, clip, width=0.95\columnwidth]{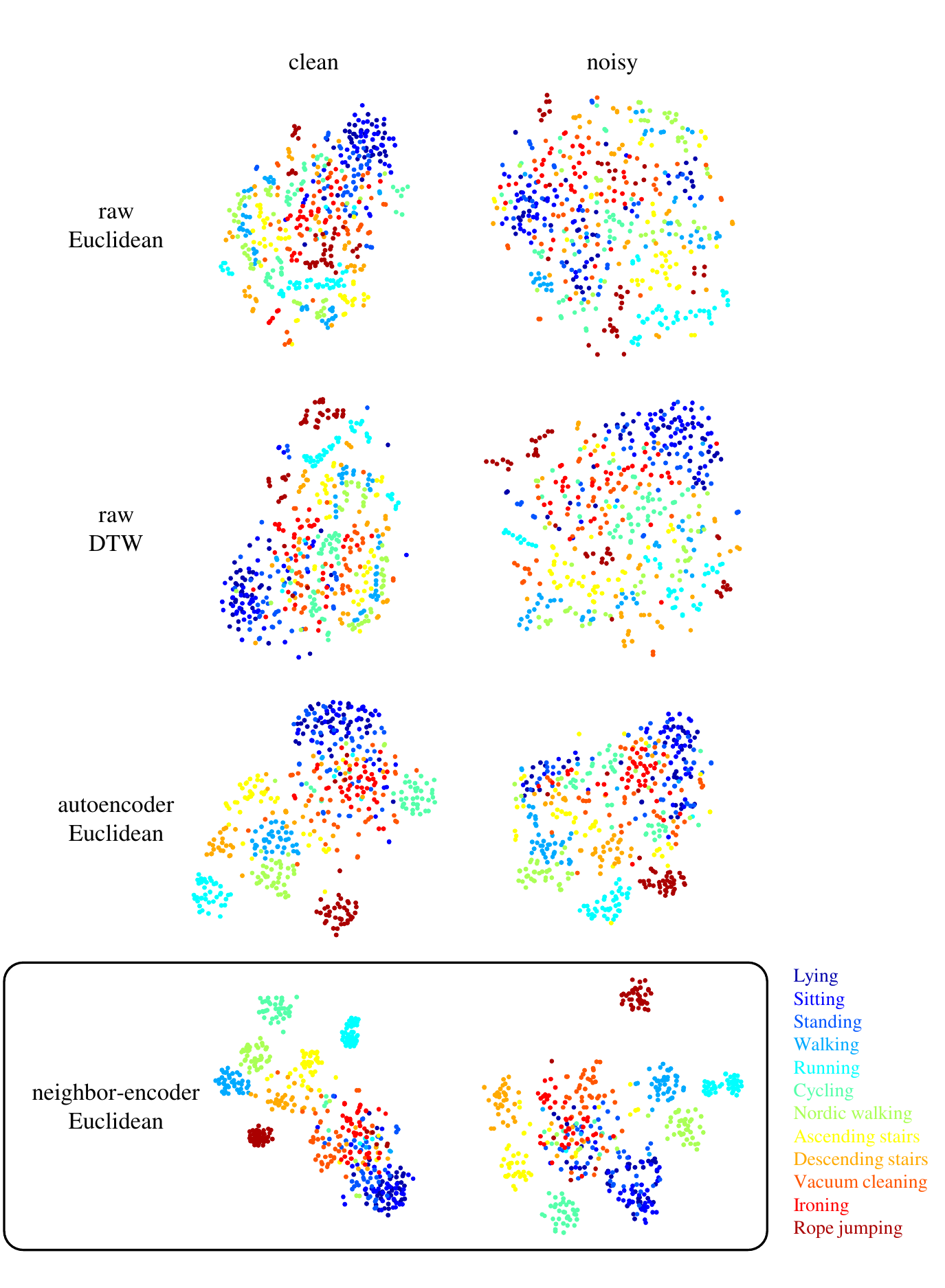}
\caption{
Visualizing the learned representation versus the raw time series on the PAMAP2 (human physical activity) data set using $t$-SNE with either Euclidean or dynamic time warping (DTW) distance~\cite{nguyen2017arxiv}.
If we manually select $27$ dimensions of the time series that are \emph{clean} and relevant (acceleration, gyroscope, magnetometer, etc.), the representation learned by both autoencoder and neighbor-encoder achieves better class separation than raw data. 
However, if the data includes \emph{noisy} and/or irrelevant dimensions (heart rate, temperature, etc.), neighbor-encoder outperforms autoencoder noticeably.
}
\label{fig_tsne_scatter}
\end{figure}

In this work, we propose an unsupervised representation learning framework (i.e., neighbor-encoder) which is \emph{general}, as it can be applied to various types of data, and \emph{versatile} since domain knowledge can be added by adopting various ``off-the-shelf'' data mining algorithms for finding neighbors.
Figure~\ref{fig_tsne_scatter} previews the $t$-SNE~\cite{maaten2008jmlr} visualization produced from a human physical activity data set (see Section~\ref{exp_activity} for details).
The embedding is generated by projecting representation learned by neighbor-encoder, autoencoder, and raw data respectively to $2D$.
By using a suitable neighbor finding algorithm, the representation learned by neighbor-encoder provides a more meaningful visualization than its rival methods.

In summary, our major contributions include:
\begin{itemize}
\item We propose a \emph{general} and \emph{versatile} framework, neighbor-encoder, which incorporates domain knowledge into unsupervised representation learning by leveraging a large family of off-the-shelf similarity search techniques.
\item We demonstrate that the \emph{performance} of the representations learned by neighbor-encoder is superior to representations learned by autoencoder.
\item We showcase the \emph{applicability} of neighbor-encoder in a diverse set of domains (i.e., handwritten digit data, text, and human physical activity data) for various data mining tasks (i.e., classification, clustering, and visualization).
\end{itemize}

To allow reproducibility, all the codes and models associated with the paper can be downloaded from~\citeauthor{nnwebsite}~(\citeyear{nnwebsite}).
The rest of this paper is organized as follows. 
In Section \ref{relatedwork} we consider related work. 
Section \ref{framework} we introduce the propose neighbor-encoder framework. 
We perform a comprehensive evaluation in Section \ref{exp} before offering conclusions and directions for future research in Section \ref{conclusion}.

\section{Related Work}
\label{relatedwork}
\textbf{Unsupervised representation learning} is usually achieved by optimizing either domain-specific objectives or general unsupervised objectives.
For example, in the domain of computer vision and music processing, unsupervised representation learning problem is formulated as a supervised learning problem with surrogate labels, generated by exploiting the temporal coherence in videos and music~\cite{agrawal2015iccv,jayaraman2015iccv,wang2015iccv,pathak2017cvpr,huang2017arxiv}. 
In the case of natural language processing, word embedding can be achieved by optimizing an objective function that ``pushes'' words occurring in a similar context (i.e., surrounded by similar words) closer in the embedding space~\cite{mikolov2013nips}.
Alternatively, general unsupervised objectives are also useful for unsupervised representation learning.
For example, minimizing the self-reconstruction error is used in autoencoder~\cite{bengio2007nips,huang2007cvpr,vincent2010jmlr}, while optimizing the $k$-means objective is shown effective in~\citeauthor{coates2012nn}~(\citeyear{coates2012nn}) and~\citeauthor{yang2017icml}~(\citeyear{yang2017icml}).
Other objectives, such as self-organizing map criteria~\cite{kohonen1982bc,bojanowski2017icml} and adversarial training~\cite{goodfellow2014nips,donahue2016arxiv,radford2015arxiv,larsen2015arxiv}, are also effective objectives for unsupervised representation learning.

\noindent\textbf{Autoencoder} is a decade-old unsupervised learning framework for dimension reduction, representation learning, and deep hierarchical model pre-training; many variants have been proposed since its initial introduction~\cite{bengio2007nips,goodfellow2016book}.
For example, the denoising autoencoder reconstructs the input data from its corrupted version; such modification improves the robustness of the learned representation~\cite{vincent2010jmlr}.
The variational autoencoder (VAE) regularizes the learning process by imposing a standard normal prior over the latent variable (i.e., representation), and such constraints help the autoencoder learn a valid generative model~\cite{kingma2013arxiv,rezende2014arxiv}.
\citeauthor{larsen2015arxiv}~(\citeyear{larsen2015arxiv}) and~\citeauthor{makhzani2015arxiv}~(\citeyear{makhzani2015arxiv}) further improves generative model learning by combining VAE with adversarial training.
Sparsity constraints on the learned representation are another form of regularization for autoencoders to learn a more discriminating representation for classification; both the $k$-sparse autoencoder~\cite{makhzani2013arxiv,makhzani2015nips} and $k$-competitive autoencoder~\cite{chen2017kdd} incorporate such ideas.

\section{Neighbor-encoder Framework}
\label{framework}
\begin{figure}[htbp]
\centering
\begin{subfigure}[b]{0.30\columnwidth}
\includegraphics[trim={14.8cm 8.25cm 14.9cm 8.55cm}, clip, width=\textwidth]{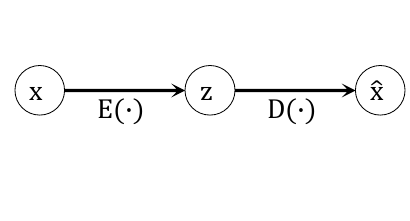}
\caption{}
\label{fig_scheme_auto}
\end{subfigure}\hskip 1em
\begin{subfigure}[b]{0.30\columnwidth}
\includegraphics[trim={14.8cm 8.25cm 14.9cm 8.55cm}, clip, width=\textwidth]{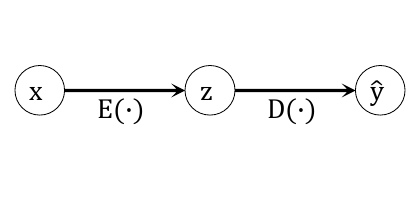}
\caption{}
\label{fig_scheme_neighbor}
\end{subfigure}\hskip 1em
\begin{subfigure}[b]{0.30\columnwidth}
\includegraphics[trim={14.8cm 8.1cm 14.9cm 8.4cm}, clip, width=\textwidth]{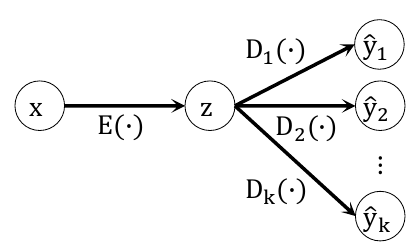}
\caption{}
\label{fig_scheme_kneighbor1}
\end{subfigure}
\caption{
Various encoder-decoder configurations for training autoencoder and neighbor-encoder:
\subref{fig_scheme_auto}) autoencoder, \subref{fig_scheme_neighbor}) neighbor-encoder, and \subref{fig_scheme_kneighbor1}) $k$-neighbor-encoder with $k$ decoders.
}
\label{fig_scheme}

\end{figure}

In this section, we introduce the proposed neighbor-encoder framework and make a comparison with autoencoder.
Figure~\ref{fig_scheme} shows different  encoder-decoder configurations for both neighbor-encoder and autoencoder.
In the following sections, we discuss the motivation and design of each encoder-decoder configuration in detail.

\subsection{Autoencoder (AE)} 
\label{framework_auto}
The overall architecture of autoencoder consists of two components: an \emph{encoder} and a \emph{decoder}.
Given input data $x$, the encoder $E(\cdot)$ is a function that encodes $x$ into a latent representation $z$ (usually in a lower dimensional space), and the decoder $D(\cdot)$ is a function that decodes $z$ in order to reconstruct $x$.
Figure~\ref{fig_scheme_auto} shows the feed-forward path of an autoencoder where $z = E(x)$ and $\hat{x} = D(z)$.
We train the autoencoder by minimizing the difference between the input data $x$ and the reconstructed data $\hat{x}$. 
Formally, given a set of training data $X$, the parameters in $E(\cdot)$ and $D(\cdot)$ are learned by minimizing the objective function $\sum_{x \in X}loss(x,\hat{x})$, where $\hat{x}=D(E(x))$.
The particular loss function we used in this work is cross entropy, but other loss function, like mean square error or mean absolute error can also be applied.
Once the autoencoder is learned, any given data can be projected to the latent representation space with $E(\cdot)$.
Both the encoder and the decoder can adopt any existing neural network architecture, such as multilayer perceptron~\cite{bengio2007nips}, convolutional net~\cite{huang2007cvpr}, or long short-term memory~\cite{hochreiter1997neural,srivastava2015icml}.

\subsection{Neighbor-encoder (NE)} 
Similar to the autoencoder, neighbor-encoder also consists of an encoder and a decoder.
Both the encoder and the decoder in neighbor-encoder work similarly to their counterparts in autoencoder; the major difference is in the objective function.
Given input data $x$ and the neighborhood function $N(\cdot)$ (which returns the neighbor $y$ of $x$), the encoder $E(\cdot)$ is a function that encodes $x$ into a latent representation $z$, and the decoder $D(\cdot)$ is a function that reconstructs $x$'s neighbor $y$ by decoding $z$.
Figure~\ref{fig_scheme_neighbor} shows the feed-forward path of a neighbor-encoder where $z = E(x)$ and $\hat{y} = D(z)$.
Formally, given a set of training data $X$ and a neighborhood function $N(\cdot)$, the neighbor-encoder is learned by minimizing the objective function $\sum_{x \in X}loss(y, \hat{y})$, where $y=N(x)$ and $\hat{y}=D(E(x))$.
Neighbor-encoder can be considered as a generalization of autoencoder as the input data can be treated as the nearest neighbor of itself with zero distance.
Note that here \emph{neighbor} can be defined in a variety of ways.
We will introduce examples of different neighbor definitions later in Section~\ref{nnfun}.

\begin{figure}[htbp]
\centering
\begin{subfigure}[b]{0.3\columnwidth}
\includegraphics[trim={10cm 6cm 12cm 5cm}, clip, width=\textwidth]{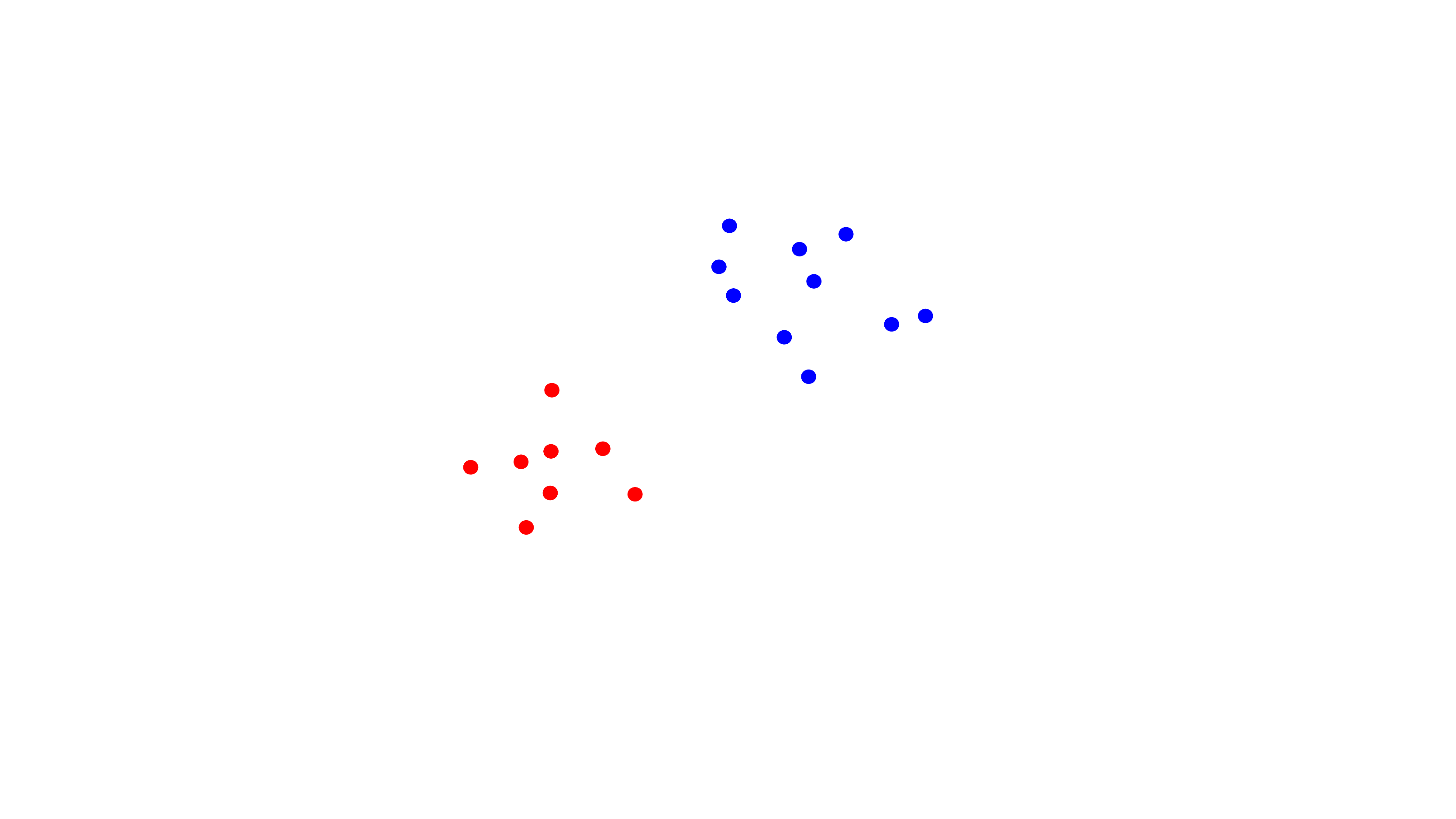}
\caption{}
\label{fig_intuition_auto}
\end{subfigure}\hskip 1em
\begin{subfigure}[b]{0.3\columnwidth}
\includegraphics[trim={10cm 6cm 12cm 5cm}, clip, width=\textwidth]{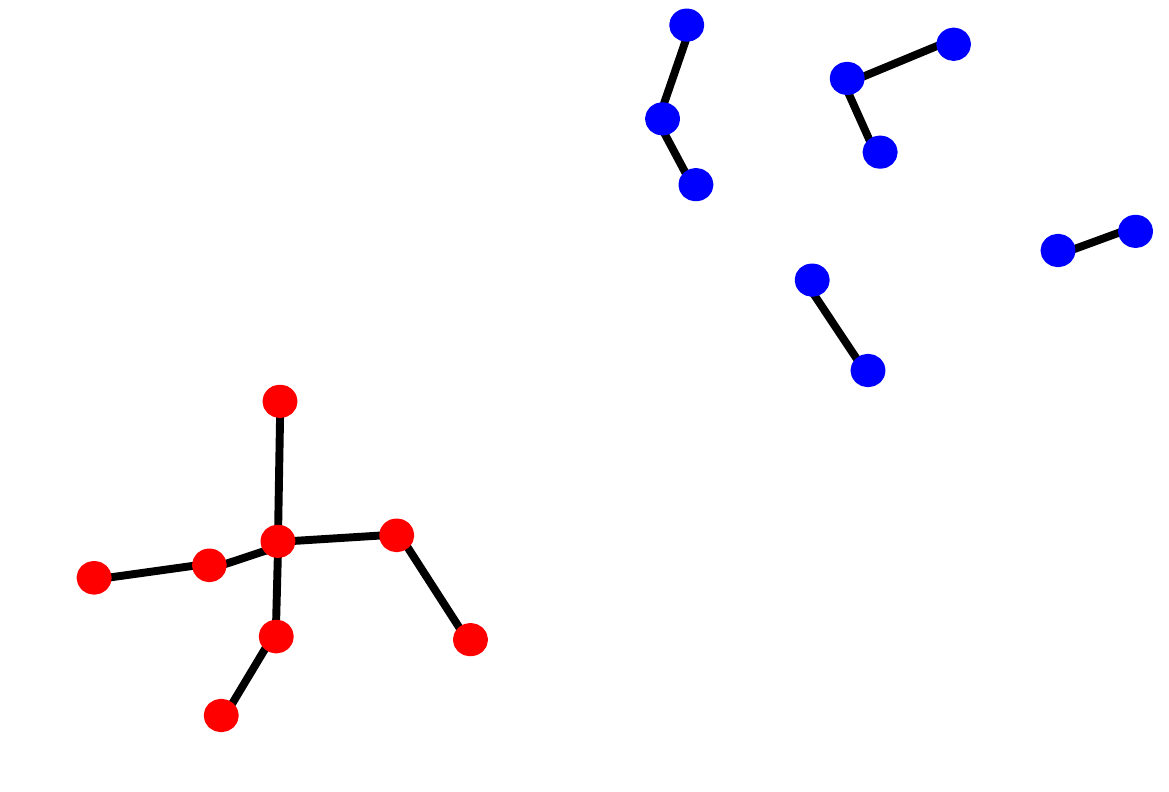}
\caption{}
\label{fig_intuition_neighbor0}
\end{subfigure}\hskip 1em
\begin{subfigure}[b]{0.3\columnwidth}
\includegraphics[trim={10cm 6cm 12cm 5cm}, clip, width=\textwidth]{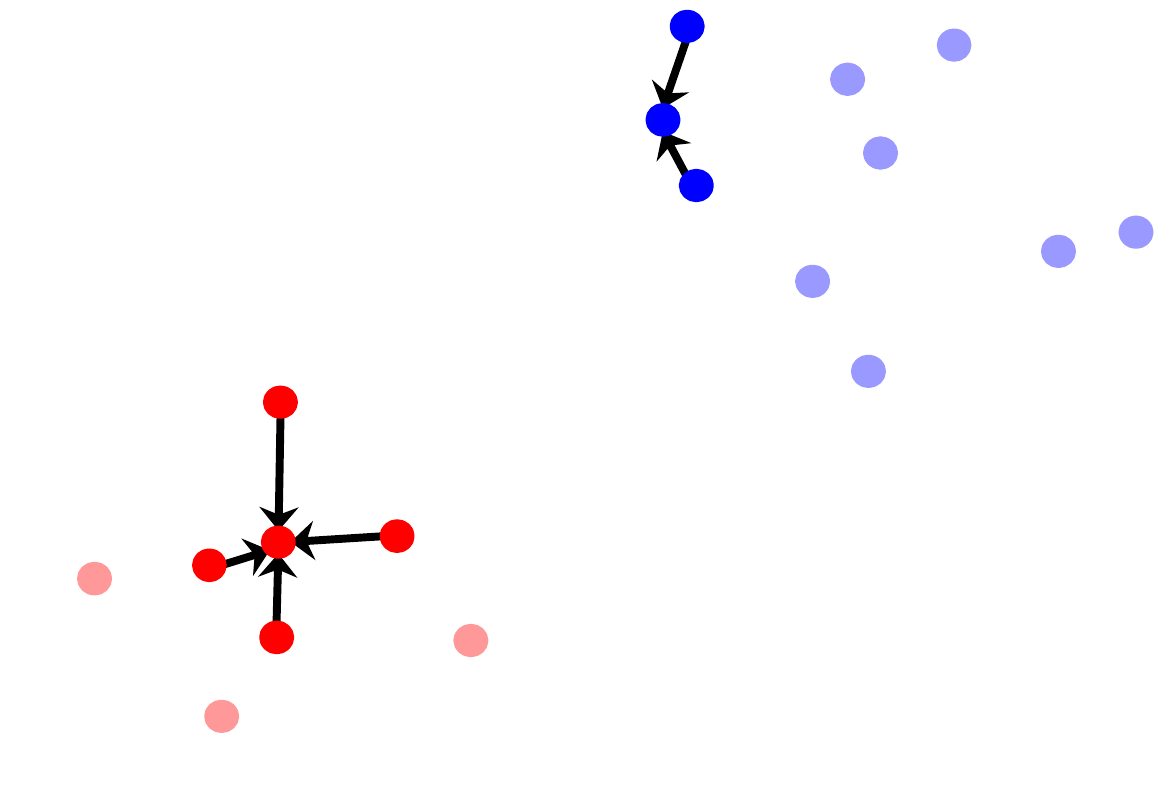}
\caption{}
\label{fig_intuition_neighbor1}
\end{subfigure}
\caption{
Intuition behind neighbor-encoder compared to autoencoder.
\subref{fig_intuition_auto}) A simple $2D$ data set with two classes, 
\subref{fig_intuition_neighbor0}) the nearest neighbor graph constructed for the data set (arrowheads are removed for clarity), and
\subref{fig_intuition_neighbor1}) an example of how neighbor-encoder would generate representation, with smaller intra-class variation for highlighted data points.
}
\label{fig_intuition_comparison}
\end{figure}

Compared to autoencoder, we argue that neighbor-encoder can better retain the similarity between data samples in the latent representation space.
Figure~\ref{fig_intuition_comparison} builds a case for this claim.
As shown in Figure~\ref{fig_intuition_auto}, we assume the data set of interest consists of samples from two classes (i.e., blue class and red class, and each class forms a cluster) in $2D$ space.
Since the autoencoder is trained by mapping each data point to itself, the learned representation for this data set would most likely be a rotated and/or re-scaled version of Figure~\ref{fig_intuition_auto}.
In contrast, the neighbor-encoder (trained with nearest neighbor relation, as shown in Figure~\ref{fig_intuition_neighbor0}) would learn a representation with much less intra-class variation.
As Figure~\ref{fig_intuition_neighbor1} shows, when several similar data points share the same nearest neighbor, the objective function will force the network to generate exactly the same output for these similar data points, thus forcing their latent representation (which is the input of the decoder) to be very similar. 

Alternatively, neighbor-encoder can be understood as a non-parametric way of generating corrupted data for denoising autoencoder.
Instead of being trained to remove arbitrary noise (e.g., Gaussian noise) from the corrupted data (which is the norm), the neighbor-encoder is trained to remove more meaningful noise from the corrupted data.
For example, a pair of nearest neighbors found using Euclidean distance in MNIST database~\cite{lecun1998ieee} usually reflects different writing styles of the same numeric digit (see Figure~\ref{fig_mnist_pair_0}).
By training the neighbor-encoder with such nearest neighbor pairs, the learning process would push the encoder network to ignore or ``remove" the writing style aspect from the handwritten digits.

Since we are using neighbor finding algorithms to guide the representation learning process, one may argue that we could instead construct a graph using the neighbor finding algorithm, then apply various graph-based representation learning methods like the ones proposed in~\cite{perozzi2014kdd,tang2015www,grover2016kdd,dong2017kdd,ribeiro2017kdd}.
Graph-based methods are indeed valid alternatives to neighbor-encoder; however, they have the following two limitations:
1) If one wishes to encode a newly obtained data, the out-of-sample problem would bring about additional complexity, as these methods are not designed to handle such a scenario. 
2) It will be impossible to learn a generative model, as graph-based methods learn the representation by modeling the relationship between examples in a data set, rather than modeling the example itself.
As a result, whenever the above limitations are crucial, the proposed neighbor-encoder is preferred over the graph-based methods.

\subsection{$k$-neighbor-encoder}
\label{framework_kne} 
Similar to the idea of generalizing the $1$-nearest neighbor classifier to a $k$-nearest neighbor classifier, neighbor-encoder can also be extended to the $k$-neighbor-encoder by reconstructing $k$ neighbors of the input data (see Figure~\ref{fig_scheme_kneighbor1}).
We train $k$ decoders to simultaneously reconstruct all $k$ neighbors of the input. 
Given an input data $x$ and the neighborhood function $N(\cdot)$ (which returns the $k$ neighbors $[y_i | \forall i \in \mathbb{Z}:0<i \leq k]$ of $x$), the encoder $E(\cdot)$ is a function that encodes $x$ into the latent representation $z$.
Then, we have a set of $k$ decoders $[D_i(\cdot) | \forall i \in \mathbb{Z}:0<i \leq k]$, in which each individual function $D_i(\cdot)$ decodes $z$ in order to reconstruct $x$'s $i$th neighbor $y_i$.

The $k$-neighbor encoder learning process is slightly more complicated than the neighbor-encoder (i.e., $1$-neighbor-encoder).
Given a set of training data $X$ and a neighborhood function $N(\cdot)$, the $k$-neighbor-encoder can be learned by minimizing $\sum_{x \in X}\sum_{y_i \in N(x)}loss(y_i, \hat{y_i})$ where $\hat{y_i}=D_i(E(x))$ and $0<i\leq k$.
Note that since there are $k$ decoders, we need to assign each $y_i$ to one of the decoders.
If there are ``naturally'' $k$ types of neighbors, we can train one decoder for each type of neighbor.
Otherwise, one possible decoder assignment strategy is choosing the decoder that provides the lowest reconstruction loss for each $y_i \in N(x)$.
This decoder assignment strategy will work if each training example has less than $k$ ``modes" of neighbors.

\subsection{Neighborhood Function}
\label{nnfun} 
To use any of the introduced neighbor-encoder configurations, we need to properly define the term neighbor.
In this section, we discuss several possible neighborhood functions for the neighbor-encoder framework.
Note that the functions listed in this section are just a small subset of all the available functions, and were chosen because they demonstrate the versatility of our approach.

\noindent\textbf{Simple Neighbor}
is defined as the objects that are closest to a given object in Euclidean distance or other distances, assuming the distance between every two objects is computable.
For example, given a set of objects $[x_1, x_2, x_3, ..., x_n]$ where each object is a real-value vector, the neighboring relationship among the objects under Euclidean distance can be approximately identified by construing a $k$-$d$ tree.

\noindent\textbf{Feature Space Neighbor}
is very similar to \emph{simple neighbor}, except that instead of computing the distance between objects in the space where the reconstruction is performed (e.g., the raw-data space), we compute the distance in an alternative representation or feature space.
To give a more concrete example, suppose we have a set of objects $[x_1, x_2, x_3, ..., x_n]$ where each object is an audio clip in mel-frequency spectrum space.
Instead of finding neighbors directly in the mel-frequency spectrum space, we transform the data into the Mel-Frequency Cepstral Coefficient (MFCC) space, as neighbors discovered in MFCC space are semantically more meaningful and searching in MFCC space is more efficient.

\noindent\textbf{Time Series Subspace Neighbor}
, as defined for multidimensional time series data, is the similarity between two objects measured by using only a subset of all dimensions.
By ignoring some dimensions, a time series could find higher quality neighbors since it is very likely that some of the dimensions contain irrelevant or noisy information (e.g., room temperature in human physical activity data).  
Given a multidimensional time series, we can use $m$STAMP~\cite{yeh2017icdm} to evaluate the neighboring relationship between all the subsequences within the time series.

\noindent\textbf{Spatial or Temporal Neighbor}
defines the neighbor based on the spatial or temporal closeness of objects. 
Specifically, given a set of objects $[x_1, x_2, x_3, ..., x_n]$ where the subscript denotes the temporal (or spatial) arrival order, $x_i$ and $x_j$ are neighbors when $|i - j| < d$, where $d$ is the predefined size of the neighborhood.
The skip-gram model in word2vec~\cite{mikolov2013nips} is an example of spatial neighbor-encoder, as the skip-gram model can be regarded as reconstructing the spatial neighbors (in the form of one-hot vector) of a given word.

\noindent\textbf{Side Information Neighbor}
defines the neighbor with side information, which could be more semantically meaningful than the aforementioned functions.
For example, images shown in the same eCommerce webpage (e.g., Amazon) would most likely belong to the same merchandise, but they can reflect different angles, colors, etc., of the merchandise.
If we select a random image from a webpage and assign it as the nearest neighbor for all the other images in the same page, we could train a representation that is invariant to view angles, lighting conditions, product variations (e.g., different color of the same smart phone), and so forth.
One may consider that using such side information implies a supervised learning system instead of an unsupervised learning system.
However, note that we only have the information regarding similar pairs while the information regarding dissimilar pairs (i.e., negative examples) is missing\footnote{
We can construct a $1$-nearest-neighbor graph by treating each image as a node and connecting each image with its nearest neighbor. 
One may sample pairs of disconnected nodes as negative examples, but such sampling method may produce false negatives, as disconnected nodes may or may not be semantically dissimilar.
}; compared to the information required by a conventional supervised learning system, this information is very limiting.

\section{Experimental Evaluation}
\label{exp}
In this section, we show the effectiveness and versatility of neighbor-encoder compared to autoencoder by performing experiments on handwritten digits, texts, and human physical activities with different neighborhood functions.
As the neighbor-encoder framework is a generalization of autoencoder, all the variants of autoencoder (e.g., denoising autoencoder~\cite{vincent2010jmlr}, variational autoencoder~\cite{kingma2013arxiv,rezende2014arxiv}, $k$-sparse autoencoder~\cite{makhzani2013arxiv,makhzani2015nips}, or adversarial autoencoder~\cite{larsen2015arxiv,makhzani2015arxiv}) can be directly ported to the neighbor-encoder framework.
As a result, we did not exhaustively test all variants of autoencoder/neighbor-encoder, but instead selected the three most popular variants (i.e., vanilla, denoising, and variational).
We leave the exhaustive comparison of the other variants for future work.

\subsection{Handwritten Digits}
\label{exp_mnist}
The MNIST database is commonly used in the initial study of newly proposed methods due to its simplicity~\cite{lecun1998ieee}.
It contains $70,000$ images of handwritten digits (one digit per image); $10,000$ of these images are test data, and the other $60,000$ are training data.
The original task for the data set is multi-class classification.
Since the proposed method is not a classifier but a representation learner (i.e., an encoder), we have evaluated our method using the following procedure: 
1) we train the encoder with all the training data, 
2) we encode both training data and test data into the learned representation space, 
3) we train a simple classifier (i.e., linear support vector machine/SVM) with various amounts of labeled training data in the representation space, then apply the classifier to the representation of test data and report the classification error (i.e.,  semi-supervised classification problem), and
4) we also apply a clustering method (i.e., $k$-means) to the representation of test data and report the adjusted Rand index.
As a proof of concept, we did not put much effort in optimizing the structure of the encoder/decoder.
We simply used a $4$-layer $2D$ convolutional net (64-64-128-128) as the encoder and a $4$-layer transposed $2D$ convolutional net (128-128-64-64) as the decoder.
We have tried several other convolutional net architectures as well; we draw the same conclusion from the experimental results with these alternative architectures.

\begin{figure}[htbp]
\centering
\includegraphics[trim={8.5cm 5.1cm 8.0cm 5.0cm}, clip, width=0.85\columnwidth]{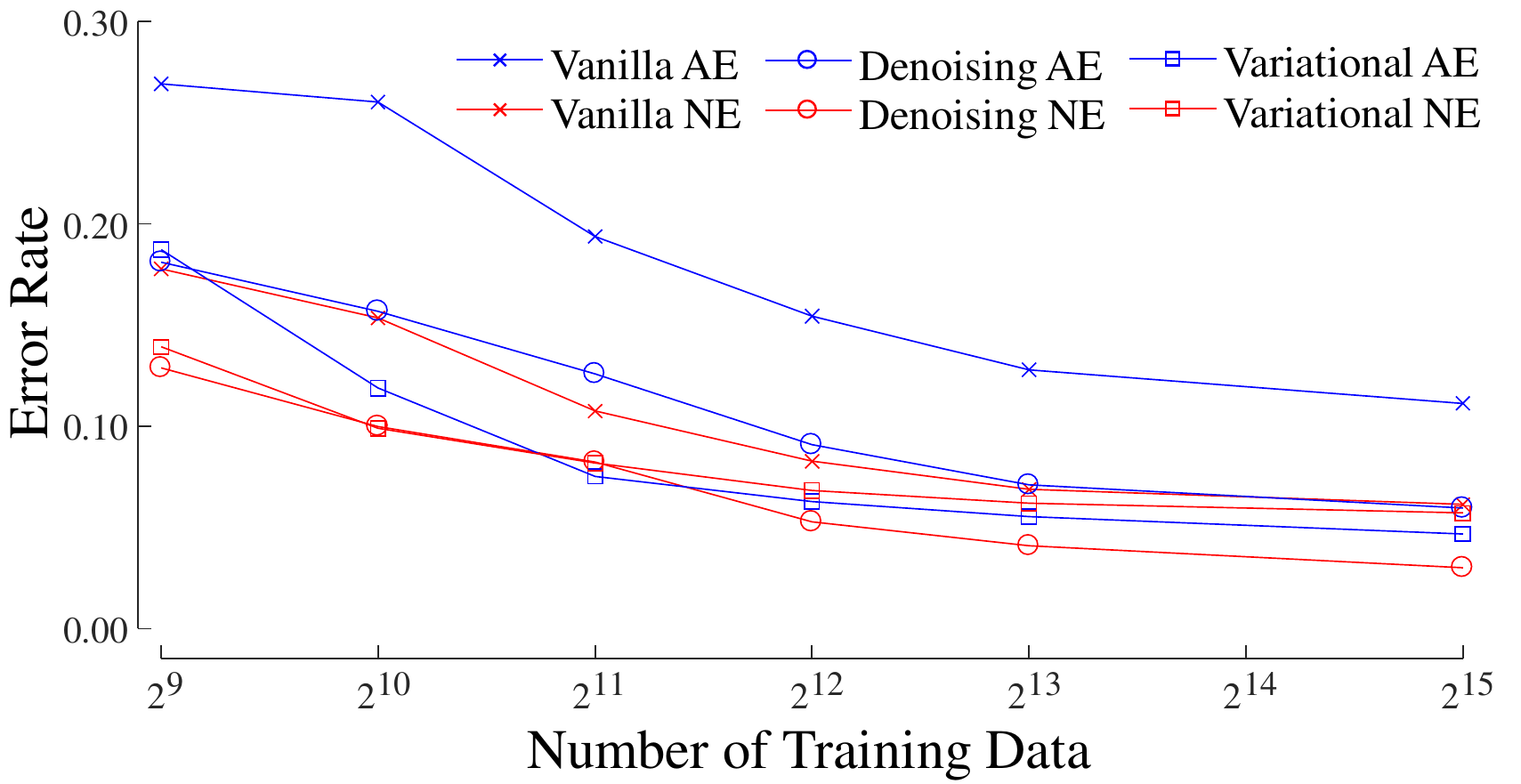}
\caption{
The classification error rate with linear SVM versus various training data sizes using different variants (i.e., vanilla, denoising, variational) of autoencoder and neighbor-encoder.
}
\label{fig_mnist_svm}
\end{figure}

Here we use the neighbor-encoder configuration (Figure~\ref{fig_scheme_neighbor}) with the simple neighbor definition for our neighbor-encoder.
We compare the performance of three variants (vanilla, denoising, and variational) of neighbor-encoder and the same three variants of autoencoder.
Figure~\ref{fig_mnist_svm} shows the classification error rate as we change the number of labeled training data for linear SVM.
All neighbor-encoder variants outperform their corresponding autoencoder variants, except the variational neighbor-encoder when the number of labeled training data is larger.
Overall, denoising neighbor-encoder produces the most discriminating representations.

Besides the semi-supervised learning experiment, we also performed a purely unsupervised clustering experiment with $k$-means.
Table~\ref{tab_mnist_kmeans} summarizes the experiment's result.
The overall conclusion is similar to that of the semi-supervised learning experiment, where all neighbor-encoder variants outperformed their corresponding autoencoder variants.
Unlike the semi-supervised experiment, variational neighbor-encoder produces the most clusterable representations in this particular experiment, but all three variants of neighbor-encoder are comparable with each other.

\begin{table}[htbp]
\centering
\caption{
The clustering adjust Rand index with $k$-means. 
}
\label{tab_mnist_kmeans}
\begin{tabular}{l|ccc}
& Vanilla & Denoising & Variational \\ \hline
AE & 0.3005 & 0.3710 & 0.4492 \\
NE & 0.4926 & 0.5039 & 0.5179 \\
\end{tabular}
\end{table}

In the previous two experiments, we define the neighbor of an object as its nearest neighbor under Euclidean distance. 
With this definition, the visual difference between an object and its neighbor is usually small, given that we have sufficient data.
To allow for more visual discrepancy between the objects and their neighbors, we could change that neighbor definition to the $i$th nearest neighbor under Euclidean distance ($i>1$).
We have repeated the clustering experiment under different settings of $i$ to examine the effect of increasing discrepancy between the objects and their neighbors.
We chose to perform the clustering experiment instead of the semi-supervised learning experiment because 1) clustering is unsupervised and 2) it is easier to present the clustering result in a single figure, as semi-supervised learning requires varying both the amount of training data and $i$.

\begin{figure}[htbp]
\centering
\begin{subfigure}[b]{0.20\columnwidth}
\includegraphics[trim={1.2cm 0.8cm 0.35cm 0.4cm}, clip, width=\textwidth]{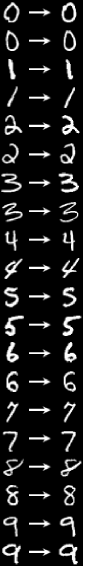}
\caption{$2^0$}
\label{fig_mnist_pair_0}
\end{subfigure}~
\begin{subfigure}[b]{0.20\columnwidth}
\includegraphics[trim={1.2cm 0.8cm 0.35cm 0.4cm}, clip, width=\textwidth]{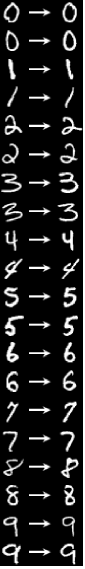}
\caption{$2^4$}
\label{fig_mnist_pair_16}
\end{subfigure}~
\begin{subfigure}[b]{0.20\columnwidth}
\includegraphics[trim={1.2cm 0.8cm 0.35cm 0.4cm}, clip, width=\textwidth]{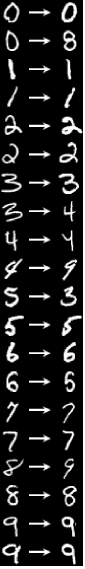}
\caption{$2^8$}
\label{fig_mnist_pair_256}
\end{subfigure}~
\begin{subfigure}[b]{0.20\columnwidth}
\includegraphics[trim={1.2cm 0.8cm 0.35cm 0.4cm}, clip, width=\textwidth]{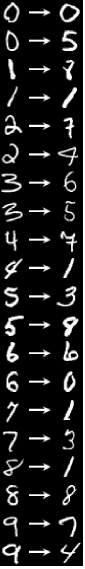}
\caption{$2^{12}$}
\label{fig_mnist_pair_4096}
\end{subfigure}
\caption{
Neighbor pairs under different proximity setting.
}
\label{fig_mnist_pair}
\end{figure}

\begin{figure}[htbp]
\centering
\includegraphics[trim={7.0cm 5.1cm 6.6cm 5.0cm}, clip, width=0.85\columnwidth]{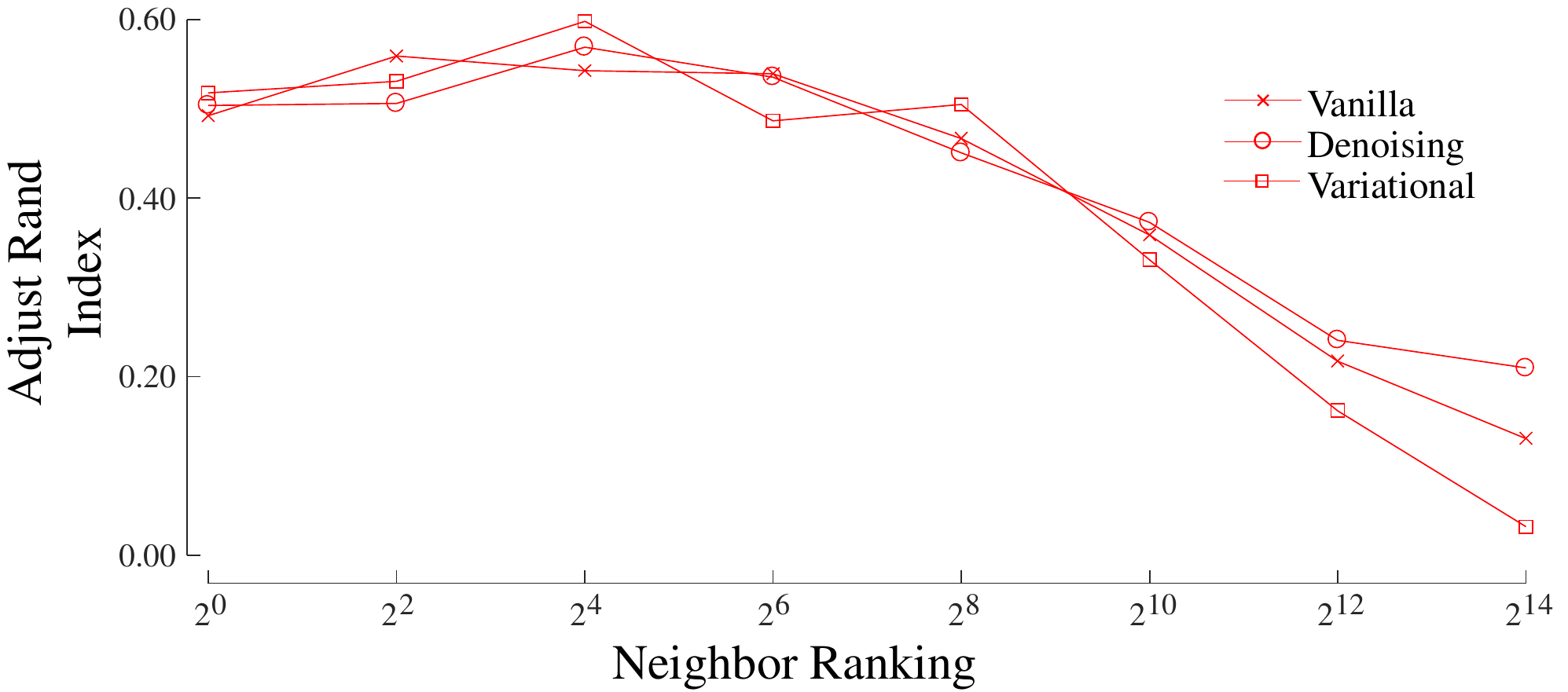}
\caption{
The clustering adjust Rand index versus the proximity of the neighbor using various neighbor-encoder variations (i.e., vanilla, denoising, variational).
The proximity of a neighbor is defined as its ranking when queried with the input.
}
\label{fig_mnist_kth_nebr}
\end{figure}

Figure~\ref{fig_mnist_kth_nebr} summarizes the result, and Figure~\ref{fig_mnist_pair} shows a randomly selected set of object-neighbor pairs under different settings of $i$.
The performance peaks around $i=2^4$ and decreases as we increase $i$; therefore, choosing the $2^4$th nearest neighbor as the reconstruction target for neighbor-encoder would create enough discrepancy between the object-neighbor pair for better representation learning.
When neighbor-encoder is used in this fashion, it can be regarded as a non-parametric way of generating noisy objects (similar as the principle of denoising autoencoder), and the settings of $i$ controls the amount of noise added to the object.
Note that neighbor-encoder is not equivalent to denoising autoencoder, as several objects can share the same $i$th nearest neighbor (recall Figure~\ref{fig_intuition_neighbor1}), but denoising autoencoder would most likely generate different noisy inputs for different objects.

\begin{figure*}[htbp]
\centering
\begin{subfigure}[b]{0.40\columnwidth}
\includegraphics[trim={1.1cm 0.8cm 0.5cm 0.4cm}, clip, width=\textwidth]{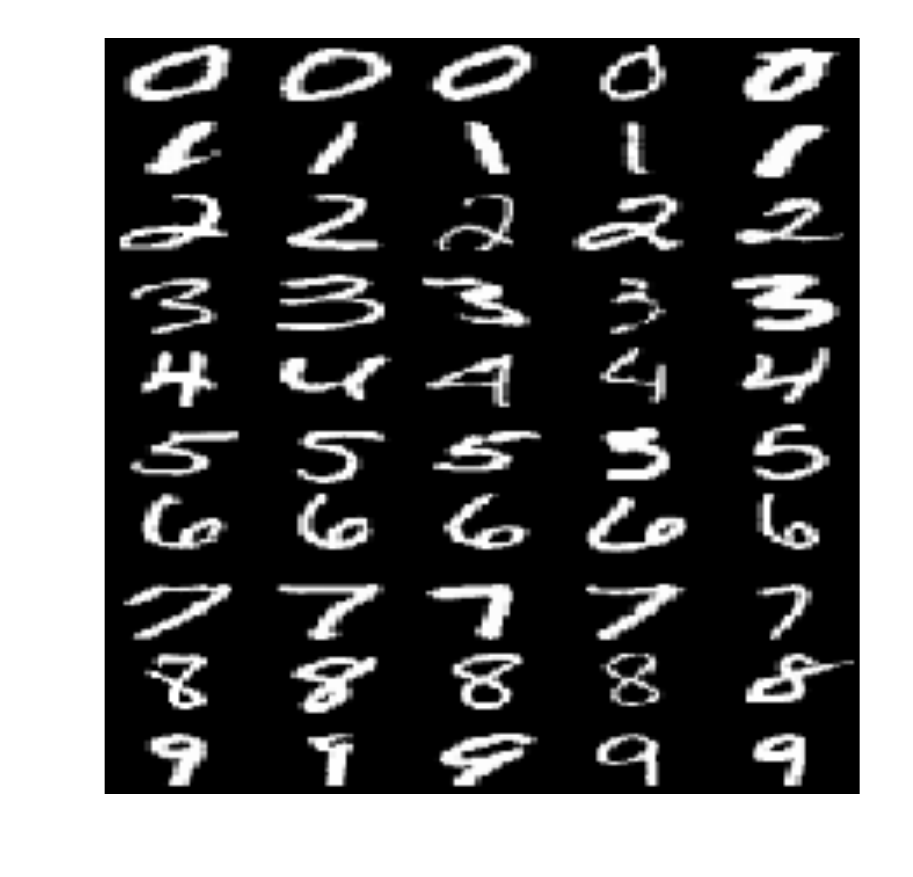}
\caption{Input}
\label{fig_mnist_input}
\end{subfigure}~~~
\begin{subfigure}[b]{0.40\columnwidth}
\includegraphics[trim={1.1cm 0.8cm 0.5cm 0.4cm}, clip, width=\textwidth]{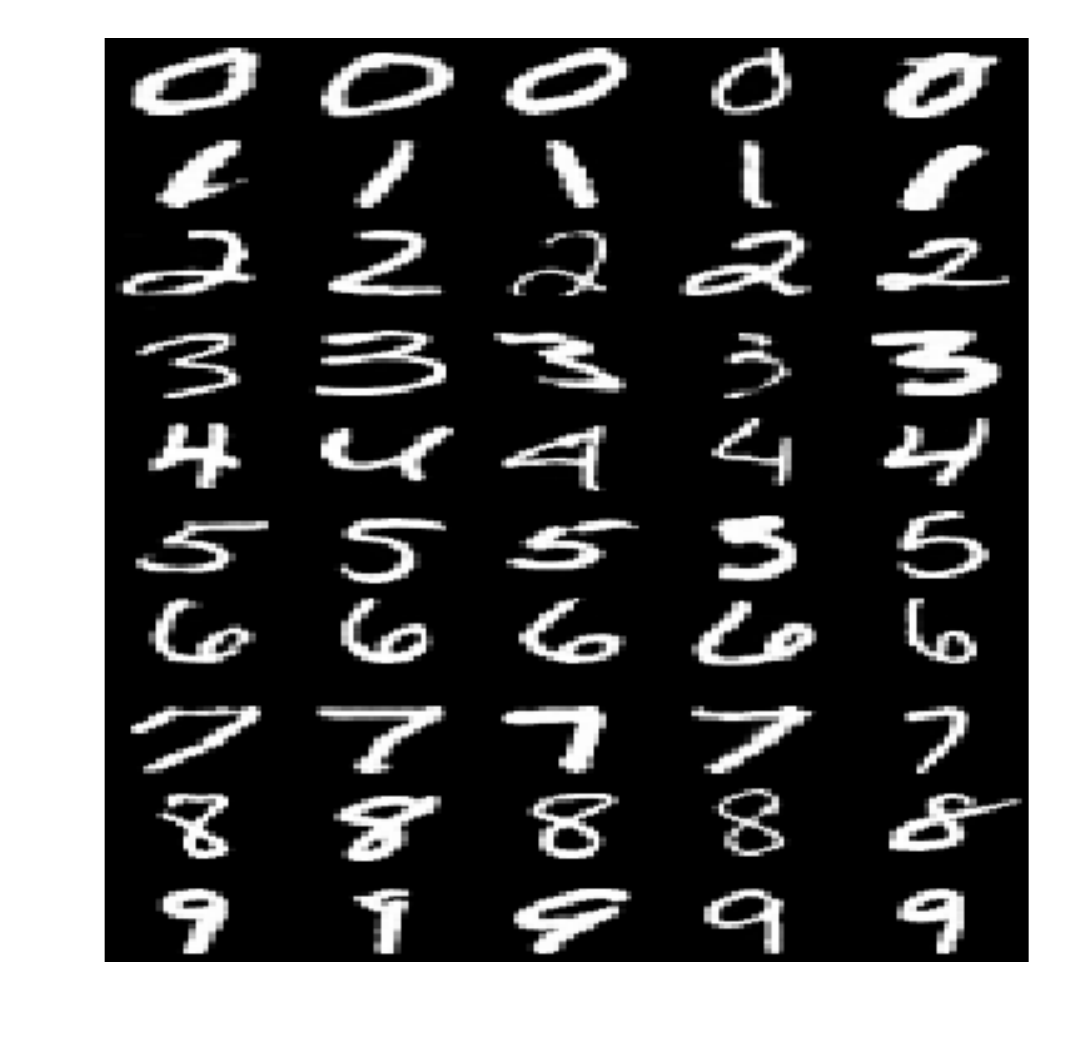}
\caption{Vanilla AE}
\label{fig_mnist_self_ae}
\end{subfigure}~
\begin{subfigure}[b]{0.40\columnwidth}
\includegraphics[trim={1.1cm 0.8cm 0.5cm 0.4cm}, clip, width=\textwidth]{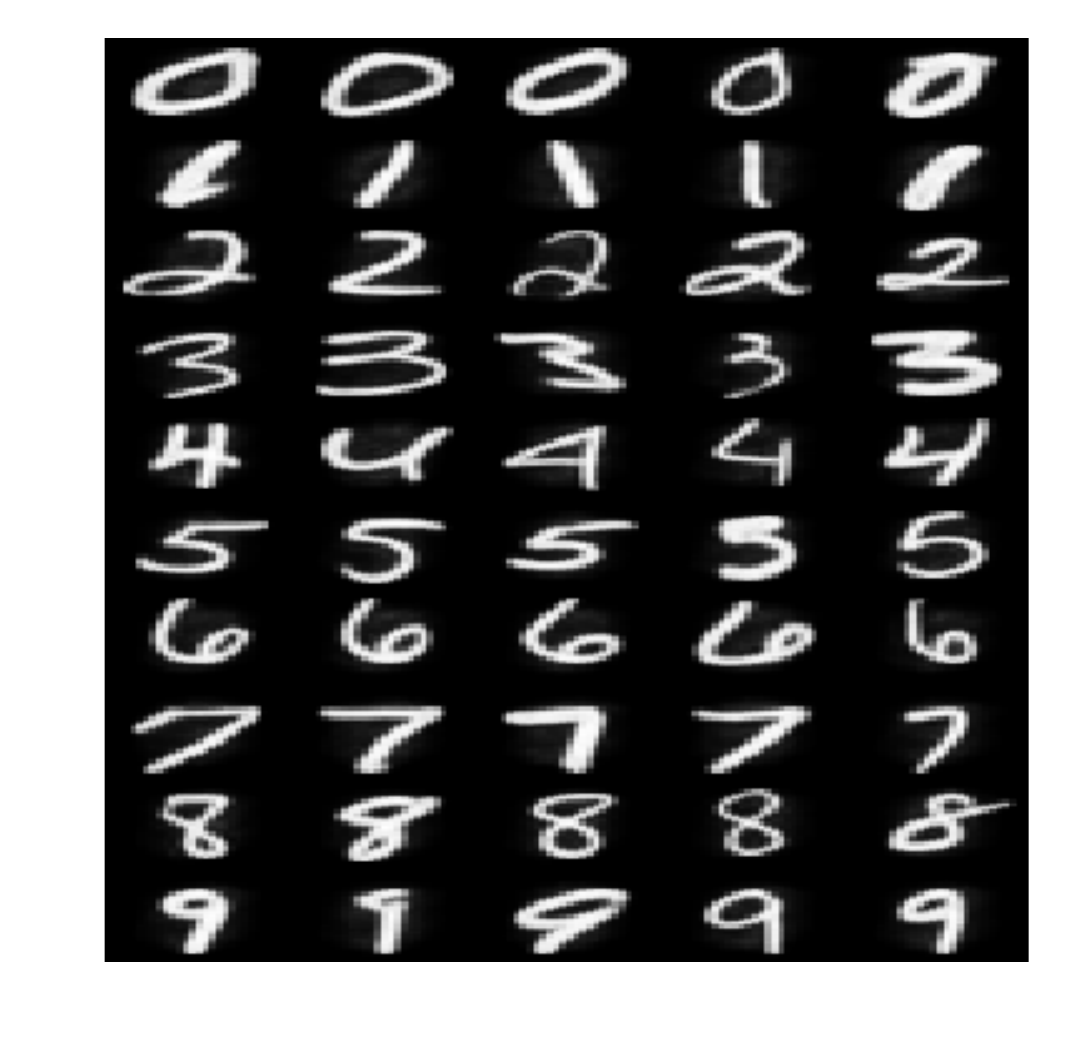}
\caption{Denoising AE}
\label{fig_mnist_self_dae}
\end{subfigure}~
\begin{subfigure}[b]{0.40\columnwidth}
\includegraphics[trim={1.1cm 0.8cm 0.5cm 0.4cm}, clip, width=\textwidth]{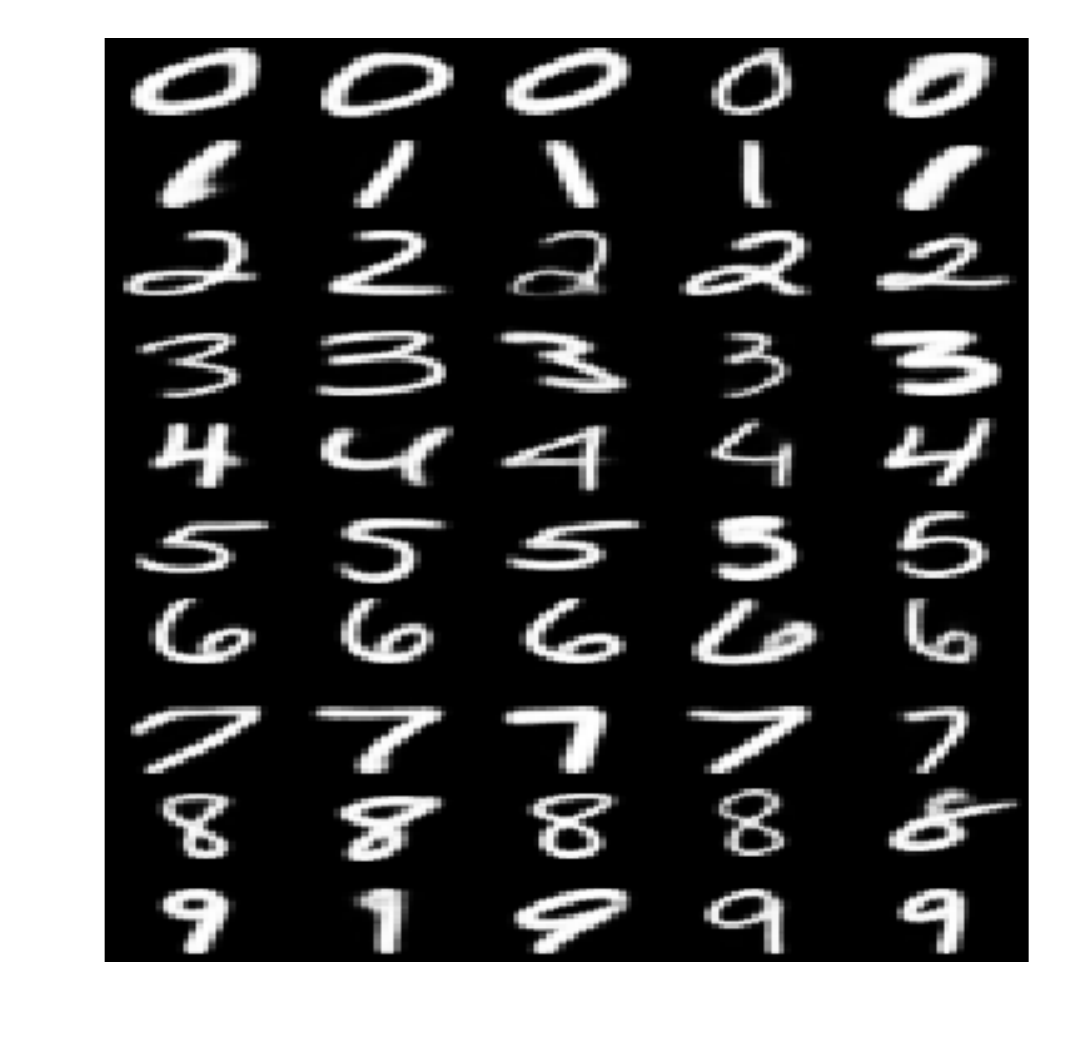}
\caption{Variational AE}
\label{fig_mnist_self_vae}
\end{subfigure}\\ \hspace{0.40\columnwidth}~~~
\begin{subfigure}[b]{0.40\columnwidth}
\includegraphics[trim={1.1cm 0.8cm 0.5cm 0.4cm}, clip, width=\textwidth]{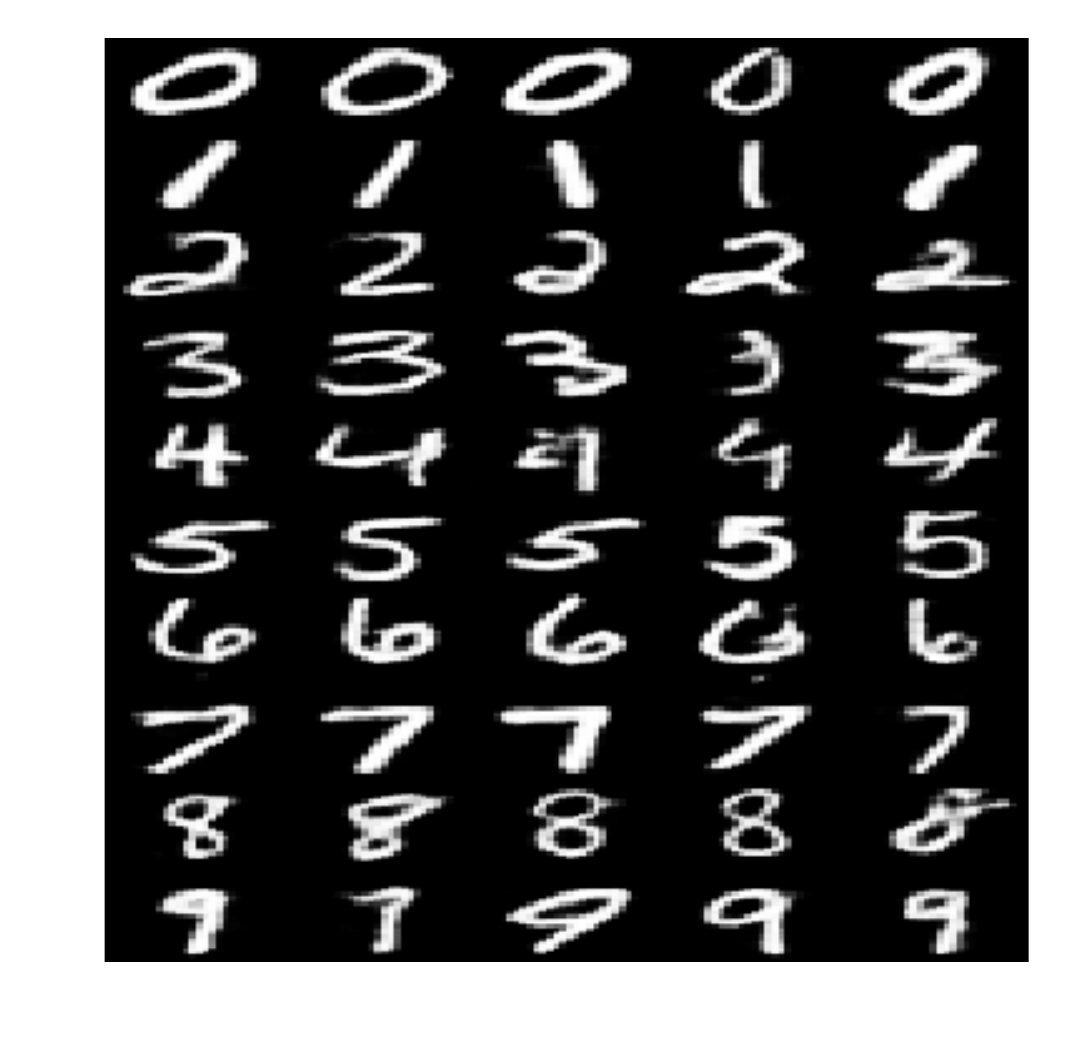}
\caption{Vanilla NE}
\label{fig_mnist_nebr_ae}
\end{subfigure}~
\begin{subfigure}[b]{0.40\columnwidth}
\includegraphics[trim={1.1cm 0.8cm 0.5cm 0.4cm}, clip, width=\textwidth]{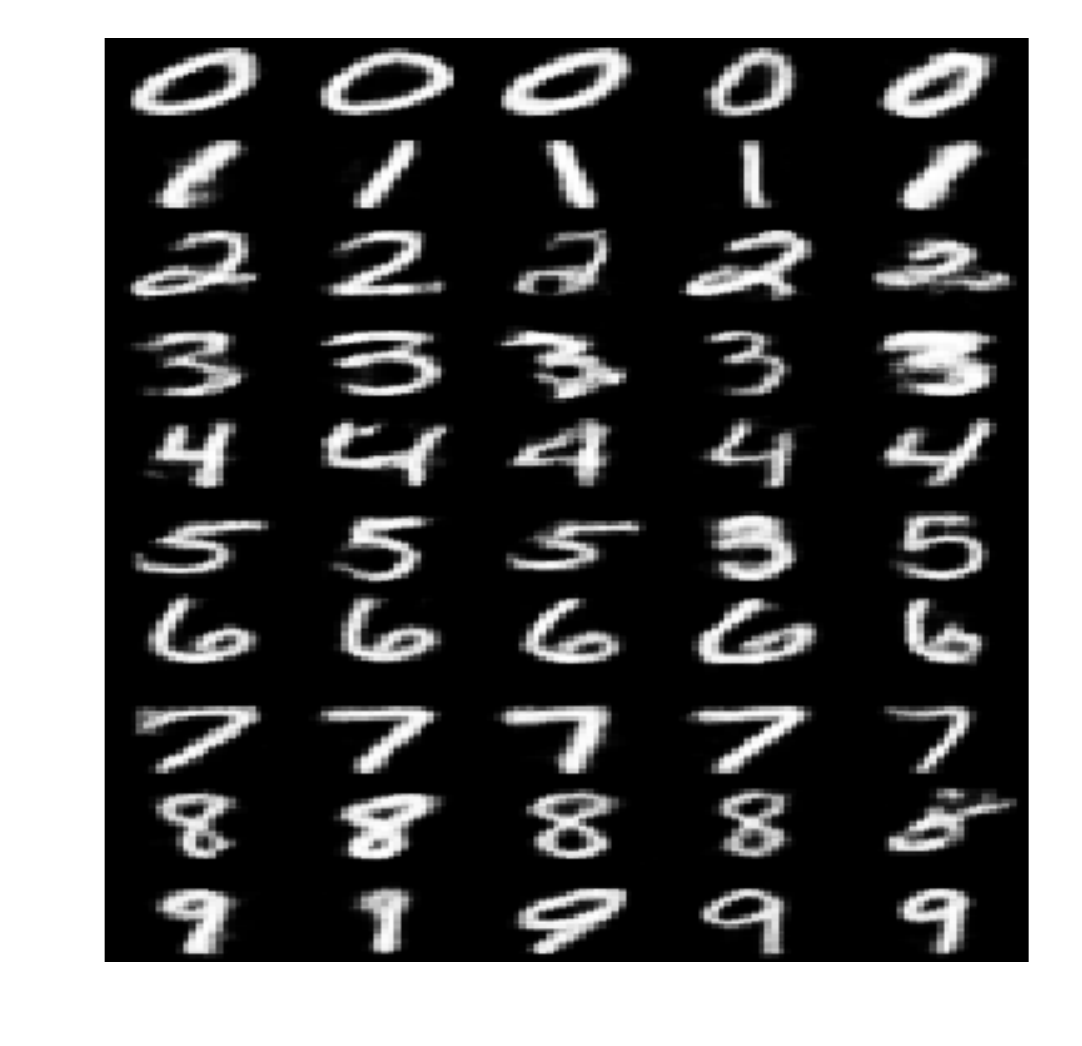}
\caption{Denoising NE}
\label{fig_mnist_nebr_dae}
\end{subfigure}~
\begin{subfigure}[b]{0.40\columnwidth}
\includegraphics[trim={1.1cm 0.8cm 0.5cm 0.4cm}, clip, width=\textwidth]{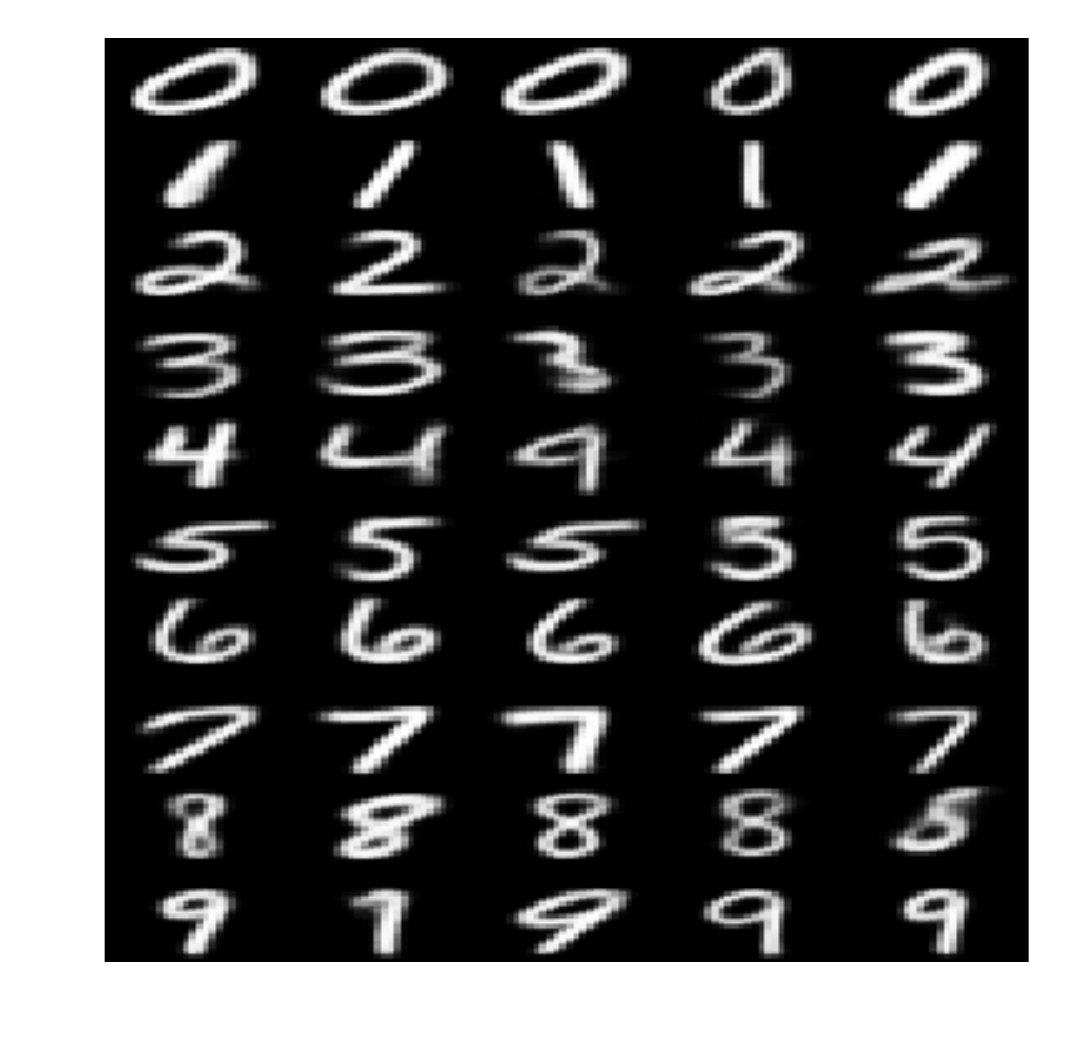}
\caption{Variational NE}
\label{fig_mnist_nebr_vae}
\end{subfigure}\\
\caption{
Outputs of the decoders for different autoencoder (AE) and neighbor-encoder (NE) variations.
}
\label{fig_mnist_recon}
\end{figure*}

To explain the performance difference between autoencoder and neighbor-encoder, we randomly selected five test examples from each class (see Figure~\ref{fig_mnist_input}) and fed them through both the autoencoder and the neighbor-encoder trained in the previous experiments.  
The outputs are shown in Figure~\ref{fig_mnist_recon}, where the top row and bottom row are autoencoder and neighbor-encoder respectively.
As expected, the output of autoencoder is almost identical to the input image.
Although the output of neighbor-encoder is still very similar to the input image, the intra-class variation is less than the output of autoencoder. 
This is because neighbor-encoder tends to reconstruct the same neighbor image from similar input data points (recall Figure~\ref{fig_intuition_neighbor1}). 
As a result, the latent representation learned by neighbor-encoder is able to achieve better performances.

\subsection{Texts}
\label{exp_texts}
The 20Newsgroup\footnote{downloaded from~\citeauthor{cardoso2007phd}~(\citeyear{cardoso2007phd})} data set contains nearly 20,000 newsgroup posts grouped into 20 (almost) balanced newsgroups/classes.
It is a popular data set for experimenting with machine learning algorithms on text documents.
We follow the clustering experiment setup presented in~\citeauthor{yang2017icml}~(\citeyear{yang2017icml}), wherein each document is represented as a tf-idf vector (using the 2,000 most frequent words in the corpus), and the performance of a method is measured by the normalized mutual information (NMI), adjusted Rand index (ARI), and clustering accuracy (ACC).
To ensure the fairness of the comparison, we use an identical network structure (250-100-20 multilayer perceptron) for the encoder~\cite{yang2017icml}.

We test three different autoencoder variants (vanilla autoencoder/AE, denoising autoencoder/DAE, and variational autoencoder/VAE) as the baselines, and enhance the best variant with the neighbor-encoder objective function (denoising neighbor-encoder/DNE).
The neighbor definition adopted in this set of experiments is the feature space neighbor, where we find the nearest neighbor of each document in the current encoding space at each epoch.
We use $k$-means (KM) to cluster the learned representation.
Table~\ref{tab_texts_20news} shows our experiment results accompanied by the experiment result reported in~\citeauthor{yang2017icml}~(\citeyear{yang2017icml}).
The proposed method (neighbor-encoder), when combined with the best variant of autoencoder, outperforms all other methods.

\begin{table}[htbp]
\centering
\begin{threeparttable}
\scriptsize
\caption{
The results of the experiment on 20Newsgroup. 
}
\label{tab_texts_20news}
\renewcommand\TPTminimum{\linewidth}
\begin{tabular}{p{0.2\linewidth}l|cccp{0.2\linewidth}}
& Methods & NMI & ARI & ACC & \\ \cline{2-5}
& JNKM\tnote{*} & 0.40 & 0.10 & 0.24 & \\
& XARY\tnote{*} & 0.19 & 0.02 & 0.18 & \\
& SC\tnote{*} & 0.40 & 0.17 & 0.34 & \\
& KM\tnote{*} & 0.41 & 0.15 & 0.30 & \\
& NMF+KM\tnote{*} & 0.39 & 0.17 & 0.33 & \\
& LCCF\tnote{*} & 0.46 & 0.17 & 0.32 & \\
& SAE+KM\tnote{*} & 0.47 & 0.28 & 0.42 & \\
& DCN\tnote{*} & 0.48 & 0.34 & 0.44 & \\ \cline{2-5}
& AE+KM & 0.44 & 0.29 & 0.43 & \\
& DAE+KM & 0.52 & 0.38 & 0.53 & \\
& VAE+KM & 0.41 & 0.18 & 0.31 & \\ \cline{2-5}
& DNE+KM & \textbf{0.56} & \textbf{0.41} & \textbf{0.57} & \\
\end{tabular}
\begin{tablenotes}
\item [\hspace{0.25\linewidth}*] Experiment results reported by~\citeauthor{yang2017icml}~(\citeyear{yang2017icml}).
\end{tablenotes}
\end{threeparttable}
\end{table}

The most similar systems (to our baselines) examined by~\citeauthor{yang2017icml}~(\citeyear{yang2017icml}) is the stacked autoencoder with $k$-means (SAE+KM).
When comparing our baselines with SAE+KM, AE+KM unsurprisingly performs similar to SAE+KM, as they are almost identical.
Out of our three baselines, the denoising autoencoder outperforms the other two variants considerably, with the variational autoencoder being the worst system. 
Because the denoising is the best autoencoder variant, we decided to extend it with the neighbor reconstruction loss function.
The resulting system (DNE+KM) outperforms all other systems, including the previous state-of-the-art deep clustering network (DCN).

Finally, we apply DNE+KM to a larger data set with imbalanced classes, RCV1-v2~\cite{lewis2004jmlr}, following the experiment/encoder setup with 20 clusters outlined in~\citeauthor{yang2017icml}~(\citeyear{yang2017icml}). 
Table~\ref{tab_texts_rcv1} summarizes the results. 
The performance of DNE+KM is similar to DCN in terms of NMI, while outperforming DCN in terms of ARI/ACC.

\begin{table}[htbp]
\centering
\begin{threeparttable}
\scriptsize
\caption{
The result of the experiment on RCV1-v2 with 20 clusters. 
}
\label{tab_texts_rcv1}
\renewcommand\TPTminimum{\linewidth}
\begin{tabular}{p{0.2\linewidth}l|cccp{0.2\linewidth}}
& Methods & NMI & ARI & ACC & \\ \cline{2-5}
& XARY\tnote{*} & 0.25 & 0.04 & 0.28 & \\
& DEC\tnote{*} & 0.08 & 0.01 & 0.14 & \\
& KM\tnote{*} & 0.58 & 0.29 & 0.47 & \\
& SAE+KM\tnote{*} & 0.59 & 0.33 & 0.46 & \\
& DCN\tnote{*} & \textbf{0.61} & 0.33 & 0.47 & \\ \cline{2-5}
& DNE+KM & 0.60 & \textbf{0.40} & \textbf{0.49} & \\
\end{tabular}
\begin{tablenotes}
\item [\hspace{0.25\linewidth}*] Experiment results reported by~\citeauthor{yang2017icml}~(\citeyear{yang2017icml}).
\end{tablenotes}
\end{threeparttable}
\end{table}

\subsection{Human Physical Activities}
\label{exp_activity}
In Section~\ref{framework}, we introduced the $k$-neighbor-encoder in addition to the neighbor-encoder.
Here we test the $k$-neighbor-encoder on the PAMAP2 data set~\cite{reiss2012abra,reiss2012iswc} using the time series subspace neighbor definition~\cite{yeh2017icdm}.
We chose the subspace neighbor definition because 
1) it addresses one of the commonly seen multidimensional time series problem scenarios (the existence of irrelevant/noisy dimensions),  
2) it is able to extract meaningful repeating patterns, and
3) it na\"{\i}vely gives multiple ``types'' of neighbors to each object.

The PAMAP2 data set was collected by mounting three inertial measurement units and a heart rate monitor on nine subjects, and recording them performing $18$ different physical activities (e.g., walking, running, playing soccer), with one session per subject, each ranging from $0.5$ hours to $1.9$ hours.
The subjects performed one activity for a few minutes, took a short break, then continued performing another activity.
In order to transfer the data set into a format that we can use for evaluation (i.e., a training/test split), 
for each subject (or recording session) we cut the data into segments according to their corresponding physical activities; 
then, within each activity segment, we generated training data from the first half, and test data from the second half with a sliding window length of $100$ and a step size of one.
We make sure that there is no overlap between training data and test data.
After the reorganization, we end up with none data sets (one pair of training/test set per subject). 
We ran experiments on each data set independently, and report averaged performance results.


The experiment procedure is very similar to the one presented in Section~\ref{exp_mnist}.
We perform the experiments under two different scenarios: ``clean" and ``noisy."
In the ``clean'' scenario, we manually deleted some dimensions of the data that are irrelevant (or harmful) to the classification/clustering tasks,
while in the ``noisy'' scenario, all dimensions of the data are retained.  
Here we use a $4$-layer $1D$ convolutional net (64-64-128-256) as the encoder, and a $4$-layer transposed $1D$ convolutional net (256-128-64-64) as the decoder.
Similar to Section~\ref{exp_mnist}, we did not put much effort in optimizing the structure of this network architecture.
We have tried modifying the convolutional net architectures in various ways, such as adding batch normalization, changing the number of layers, or varying the number of filters for each layer, etc., and the conclusion drawn from the experimental results remains virtually unchanged. 

\begin{figure}[htbp]
\begin{subfigure}[b]{0.85\columnwidth}
\centering
\includegraphics[trim={9.3cm 5.5cm 8.0cm 5.0cm}, clip, width=\textwidth]{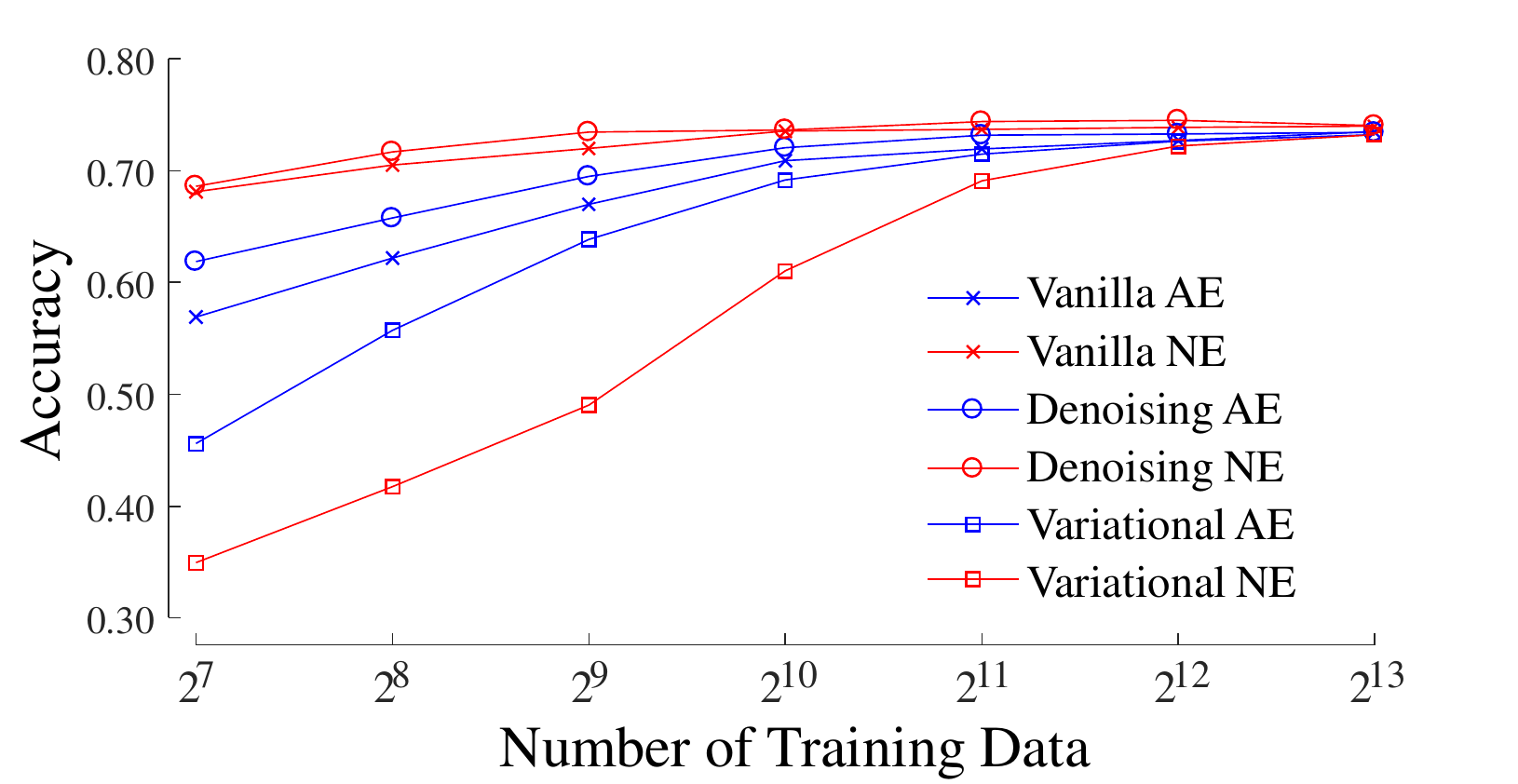}
\caption{Clean scenario}
\label{fig_pamap2_svm_clean}
\end{subfigure} \\
\begin{subfigure}[b]{0.85\columnwidth}
\centering
\includegraphics[trim={9.3cm 5.5cm 8.0cm 5.0cm}, clip, width=\textwidth]{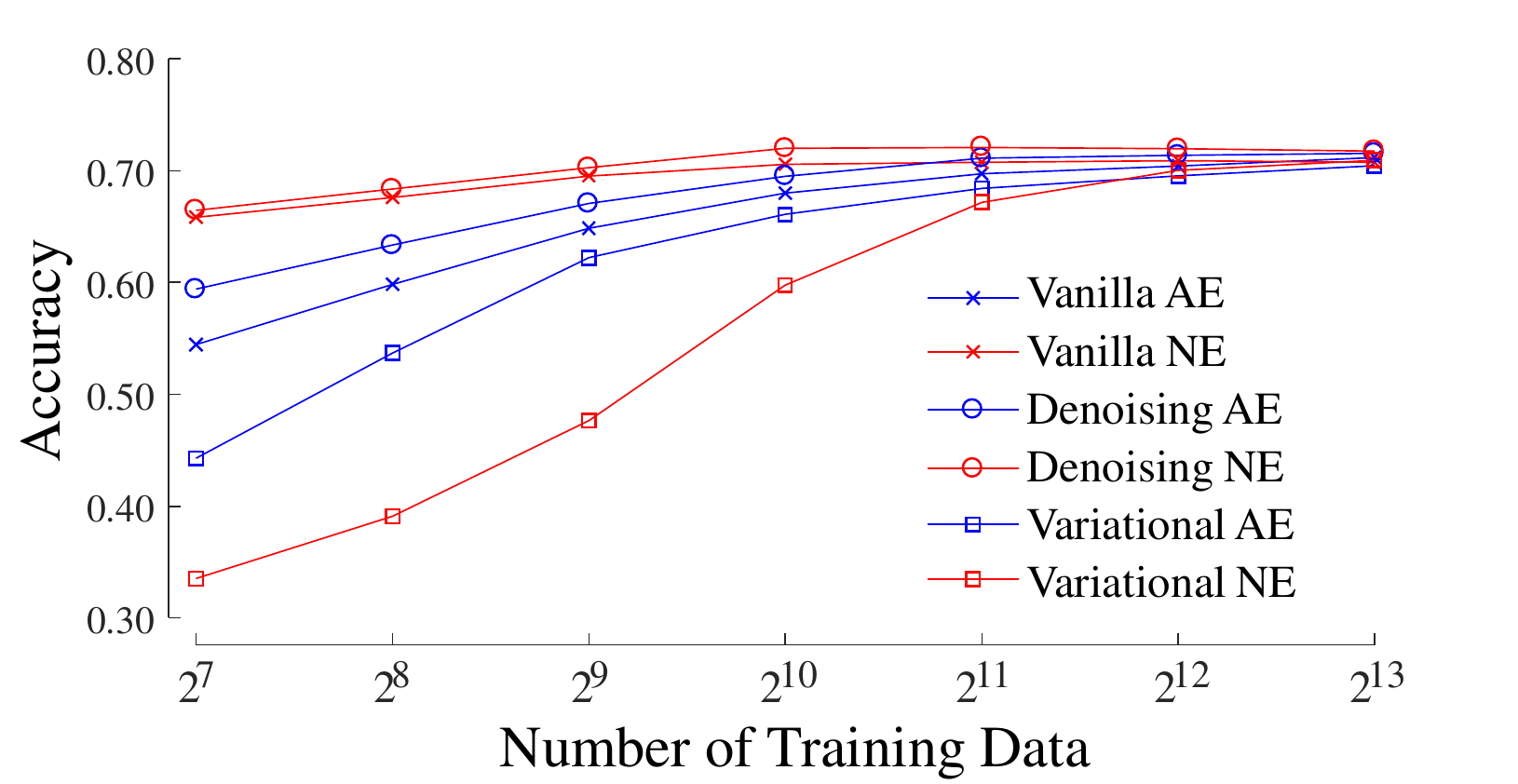}
\caption{Noisy scenario}
\label{fig_pamap2_svm_noisy}
\end{subfigure}
\caption{
The classification accuracy with linear SVM versus various labeled training data size using different variants (i.e., vanilla, denoising, variational) of either autoencoder and $k$-neighbor-encoder.
}
\label{fig_pamap2_svm}
\end{figure}

In Figure~\ref{fig_pamap2_svm}, we compare the semi-supervised classification capability of vanilla, denoising, and variational autoencoder/$k$-neighbor-encoder under both the``clean" scenario and the ``noisy" scenario.
Both vanilla and denoising $k$-neighbor-encoder outperforms their corresponding autoencoder in all scenarios.
The performance difference is more notable when the number of training data is small.
On the contrary, variational autoencoder outperforms the corresponding $k$-neighbor-encoder; however, the performance of both variational autoencoder and $k$-neighbor-encoder are considerably worse than their vanilla and denoising counterparts.
Overall, both the vanilla and denoising $k$-neighbor-encoders work relatively well for this problem.

Table~\ref{tab_pamap2_kmeans} shows the clustering experiment with $k$-means.
For the vanilla encoder-decoder system, $k$-neighbor-encoder surpasses autoencoder in both scenarios, especially in the noisy scenario.
When the denoising mechanism is added to the encoder-decoder system, it greatly boosts the performance of autoencoders, but the performance of $k$-neighbor-encoder still greatly exceeds autoencoder.  
Similar to the semi-supervised learning experiment, the variational encoder-decoder system performs poorly for this data set.
In general, both the vanilla and denoising $k$-neighbor-encoders outperform their autoencoder counterparts for the clustering problem on PAMAP2 data set.

\begin{table}[htbp]
\centering
\caption{
The clustering adjust Rand index with $k$-means.
}
\label{tab_pamap2_kmeans}
\begin{tabular}{l|l|ccc}
& & Vanilla & Denoising & Variational \\ \hline
\multirow{2}{*}{Clean} & AE & 0.3815 & 0.4159 & 0.1597 \\
& NE & 0.4203 & 0.4272 & 0.1192 \\ \hline
\multirow{2}{*}{Noisy} & AE & 0.1844 & 0.2336 & 0.1034 \\
& NE & 0.3832 & 0.3948 & 0.1081 \\
\end{tabular}
\end{table}

Figure~\ref{fig_tsne_scatter} further demonstrates the advantage of neighbor-encoder over autoencoder. 
Here we use $t$-SNE to project various representations of the data of subject $1$ into $2D$ space. 
The representations include the raw data itself, the latent representation learned by denoising autoencoder, and the latent representation learned by denoising $k$-neighbor-encoder. 
Despite the clustering experiment suggests that autoencoder is comparable with $k$-neighbor-encoder, we can see that the latent representation learned by $k$-neighbor-encoder provides a much more meaningful visualization of different classes than the rival methods do (includes autoencoder) in the face of noisy/irrelevant dimensions.

\section{Conclusion}
\label{conclusion}
In this work, we have proposed an unsupervised learning framework called neighbor-encoder that is both \emph{general}, in that it can easily be applied to data in various domains, and \emph{versatile} as it can incorporate domain knowledge by utilizing different neighborhood functions. 
We have showcased the effectiveness of neighbor-encoder compared to autoencoder in various domains, including images, text, time series, and so forth. 
In future work, we plan to either
1) explore the possibility of applying neighbor-encoder to problems like one-shot learning or 
2) demonstrate the usefulness of the neighbor-encoder in more practical and applied tasks, including information retrieval.
We made all the codes/models available at~\citeauthor{nnwebsite}~(\citeyear{nnwebsite}), to allow others to confirm and expand our work.

\small{\bibliography{reference_mike}}

\begin{thebibliography}{}

\bibitem[\protect\citeauthoryear{Agrawal, Carreira, and
  Malik}{2015}]{agrawal2015iccv}
Agrawal, P.; Carreira, J.; and Malik, J.
\newblock 2015.
\newblock Learning to see by moving.
\newblock In {\em Proceedings of the IEEE International Conference on Computer
  Vision}.

\bibitem[\protect\citeauthoryear{Bengio \bgroup et al\mbox.\egroup
  }{2007}]{bengio2007nips}
Bengio, Y.; Lamblin, P.; Popovici, D.; and Larochelle, H.
\newblock 2007.
\newblock Greedy layer-wise training of deep networks.
\newblock In {\em Advances in neural information processing systems},
  153--160.

\bibitem[\protect\citeauthoryear{Bojanowski and
  Joulin}{2017}]{bojanowski2017icml}
Bojanowski, P., and Joulin, A.
\newblock 2017.
\newblock Unsupervised learning by predicting noise.
\newblock In {\em Proceedings of the 34th international conference on Machine
  learning}.

\bibitem[\protect\citeauthoryear{Cardoso-Cachopo}{2007}]{cardoso2007phd}
Cardoso-Cachopo, A.
\newblock 2007.
\newblock {Improving Methods for Single-label Text Categorization}.
\newblock PdD Thesis, Instituto Superior Tecnico, Universidade Tecnica de
  Lisboa.
\newblock \url{http://ana.cachopo.org/}.

\bibitem[\protect\citeauthoryear{Catherine and
  Cohen}{2017}]{catherine2017transnets}
Catherine, R., and Cohen, W.
\newblock 2017.
\newblock Transnets: Learning to transform for recommendation.
\newblock {\em arXiv preprint arXiv:1704.02298}.

\bibitem[\protect\citeauthoryear{Chen and Zaki}{2017}]{chen2017kdd}
Chen, Y., and Zaki, M.~J.
\newblock 2017.
\newblock Kate: K-competitive autoencoder for text.
\newblock In {\em Proceedings of the 23rd ACM SIGKDD International Conference
  on Knowledge Discovery and Data Mining},  85--94.
\newblock ACM.

\bibitem[\protect\citeauthoryear{Coates and Ng}{2012}]{coates2012nn}
Coates, A., and Ng, A.~Y.
\newblock 2012.
\newblock Learning feature representations with k-means.
\newblock In {\em Neural networks: Tricks of the trade}. Springer.
\newblock  561--580.

\bibitem[\protect\citeauthoryear{Donahue, Kr{\"a}henb{\"u}hl, and
  Darrell}{2016}]{donahue2016arxiv}
Donahue, J.; Kr{\"a}henb{\"u}hl, P.; and Darrell, T.
\newblock 2016.
\newblock Adversarial feature learning.
\newblock {\em arXiv preprint arXiv:1605.09782}.

\bibitem[\protect\citeauthoryear{Dong, Chawla, and Swami}{2017}]{dong2017kdd}
Dong, Y.; Chawla, N.~V.; and Swami, A.
\newblock 2017.
\newblock metapath2vec: Scalable representation learning for heterogeneous
  networks.
\newblock In {\em Proceedings of the 23rd ACM SIGKDD International Conference
  on Knowledge Discovery and Data Mining},  135--144.
\newblock ACM.

\bibitem[\protect\citeauthoryear{Goodfellow, Bengio, and
  Courville}{2016}]{goodfellow2016book}
Goodfellow, I.; Bengio, Y.; and Courville, A.
\newblock 2016.
\newblock {\em Deep Learning}.
\newblock MIT Press.
\newblock \url{http://www.deeplearningbook.org}.

\bibitem[\protect\citeauthoryear{Goodfellow \bgroup et al\mbox.\egroup
  }{2014}]{goodfellow2014nips}
Goodfellow, I.; Pouget-Abadie, J.; Mirza, M.; Xu, B.; Warde-Farley, D.; Ozair,
  S.; Courville, A.; and Bengio, Y.
\newblock 2014.
\newblock Generative adversarial nets.
\newblock In {\em Advances in neural information processing systems},
  2672--2680.

\bibitem[\protect\citeauthoryear{Grover and Leskovec}{2016}]{grover2016kdd}
Grover, A., and Leskovec, J.
\newblock 2016.
\newblock node2vec: Scalable feature learning for networks.
\newblock In {\em Proceedings of the 22nd ACM SIGKDD international conference
  on Knowledge discovery and data mining},  855--864.
\newblock ACM.

\bibitem[\protect\citeauthoryear{Hochreiter and
  Schmidhuber}{1997}]{hochreiter1997neural}
Hochreiter, S., and Schmidhuber, J.
\newblock 1997.
\newblock Long short-term memory.
\newblock {\em Neural computation} 9(8):1735--1780.

\bibitem[\protect\citeauthoryear{Huang \bgroup et al\mbox.\egroup
  }{2007}]{huang2007cvpr}
Huang, F.~J.; Boureau, Y.-L.; LeCun, Y.; et~al.
\newblock 2007.
\newblock Unsupervised learning of invariant feature hierarchies with
  applications to object recognition.
\newblock In {\em Computer Vision and Pattern Recognition, 2007. IEEE
  Conference on},  1--8.
\newblock IEEE.

\bibitem[\protect\citeauthoryear{Huang, Chou, and Yang}{2017}]{huang2017arxiv}
Huang, Y.-S.; Chou, S.-Y.; and Yang, Y.-H.
\newblock 2017.
\newblock Similarity embedding network for unsupervised sequential pattern
  learning by playing music puzzle games.
\newblock {\em arXiv preprint arXiv:1709.04384}.

\bibitem[\protect\citeauthoryear{Jayaraman and
  Grauman}{2015}]{jayaraman2015iccv}
Jayaraman, D., and Grauman, K.
\newblock 2015.
\newblock Learning image representations tied to ego-motion.
\newblock In {\em Proceedings of the IEEE International Conference on Computer
  Vision},  1413--1421.

\bibitem[\protect\citeauthoryear{Kingma and Welling}{2013}]{kingma2013arxiv}
Kingma, D.~P., and Welling, M.
\newblock 2013.
\newblock Auto-encoding variational bayes.
\newblock {\em arXiv preprint arXiv:1312.6114}.

\bibitem[\protect\citeauthoryear{Kohonen}{1982}]{kohonen1982bc}
Kohonen, T.
\newblock 1982.
\newblock Self-organized formation of topologically correct feature maps.
\newblock {\em Biological cybernetics} 43(1):59--69.

\bibitem[\protect\citeauthoryear{Larsen \bgroup et al\mbox.\egroup
  }{2015}]{larsen2015arxiv}
Larsen, A. B.~L.; S{\o}nderby, S.~K.; Larochelle, H.; and Winther, O.
\newblock 2015.
\newblock Autoencoding beyond pixels using a learned similarity metric.
\newblock {\em arXiv preprint arXiv:1512.09300}.

\bibitem[\protect\citeauthoryear{LeCun \bgroup et al\mbox.\egroup
  }{1998}]{lecun1998ieee}
LeCun, Y.; Bottou, L.; Bengio, Y.; and Haffner, P.
\newblock 1998.
\newblock Gradient-based learning applied to document recognition.
\newblock {\em Proceedings of the IEEE} 86(11):2278--2324.

\bibitem[\protect\citeauthoryear{Lewis \bgroup et al\mbox.\egroup
  }{2004}]{lewis2004jmlr}
Lewis, D.~D.; Yang, Y.; Rose, T.~G.; and Li, F.
\newblock 2004.
\newblock Rcv1: A new benchmark collection for text categorization research.
\newblock {\em Journal of machine learning research} 5(Apr):361--397.

\bibitem[\protect\citeauthoryear{Lloyd}{1982}]{lloyd1982tit}
Lloyd, S.
\newblock 1982.
\newblock Least squares quantization in pcm.
\newblock {\em IEEE transactions on information theory} 28(2):129--137.

\bibitem[\protect\citeauthoryear{Maaten and Hinton}{2008}]{maaten2008jmlr}
Maaten, L. v.~d., and Hinton, G.
\newblock 2008.
\newblock Visualizing data using t-sne.
\newblock {\em Journal of Machine Learning Research} 9(Nov):2579--2605.

\bibitem[\protect\citeauthoryear{Makhzani and Frey}{2013}]{makhzani2013arxiv}
Makhzani, A., and Frey, B.
\newblock 2013.
\newblock K-sparse autoencoders.
\newblock {\em arXiv preprint arXiv:1312.5663}.

\bibitem[\protect\citeauthoryear{Makhzani and Frey}{2015}]{makhzani2015nips}
Makhzani, A., and Frey, B.~J.
\newblock 2015.
\newblock Winner-take-all autoencoders.
\newblock In {\em Advances in Neural Information Processing Systems},
  2791--2799.

\bibitem[\protect\citeauthoryear{Makhzani \bgroup et al\mbox.\egroup
  }{2015}]{makhzani2015arxiv}
Makhzani, A.; Shlens, J.; Jaitly, N.; Goodfellow, I.; and Frey, B.
\newblock 2015.
\newblock Adversarial autoencoders.
\newblock {\em arXiv preprint arXiv:1511.05644}.

\bibitem[\protect\citeauthoryear{Mikolov \bgroup et al\mbox.\egroup
  }{2013}]{mikolov2013nips}
Mikolov, T.; Sutskever, I.; Chen, K.; Corrado, G.~S.; and Dean, J.
\newblock 2013.
\newblock Distributed representations of words and phrases and their
  compositionality.
\newblock In {\em Advances in neural information processing systems},
  3111--3119.

\bibitem[\protect\citeauthoryear{Nguyen \bgroup et al\mbox.\egroup
  }{2017}]{nguyen2017arxiv}
Nguyen, M.; Purushotham, S.; To, H.; and Shahabi, C.
\newblock 2017.
\newblock m-tsne: A framework for visualizing high-dimensional multivariate
  time series.
\newblock {\em arXiv preprint arXiv:1708.07942}.

\bibitem[\protect\citeauthoryear{Pathak \bgroup et al\mbox.\egroup
  }{2017}]{pathak2017cvpr}
Pathak, D.; Girshick, R.; Doll{\'a}r, P.; Darrell, T.; and Hariharan, B.
\newblock 2017.
\newblock Learning features by watching objects move.
\newblock In {\em Computer Vision and Pattern Recognition, 2017. IEEE
  Conference on}.

\bibitem[\protect\citeauthoryear{Perozzi, Al-Rfou, and
  Skiena}{2014}]{perozzi2014kdd}
Perozzi, B.; Al-Rfou, R.; and Skiena, S.
\newblock 2014.
\newblock Deepwalk: Online learning of social representations.
\newblock In {\em Proceedings of the 20th ACM SIGKDD international conference
  on Knowledge discovery and data mining},  701--710.
\newblock ACM.

\bibitem[\protect\citeauthoryear{Radford, Metz, and
  Chintala}{2015}]{radford2015arxiv}
Radford, A.; Metz, L.; and Chintala, S.
\newblock 2015.
\newblock Unsupervised representation learning with deep convolutional
  generative adversarial networks.
\newblock {\em arXiv preprint arXiv:1511.06434}.

\bibitem[\protect\citeauthoryear{Reiss and Stricker}{2012a}]{reiss2012abra}
Reiss, A., and Stricker, D.
\newblock 2012a.
\newblock Creating and benchmarking a new dataset for physical activity
  monitoring.
\newblock In {\em Proceedings of the 5th International Conference on PErvasive
  Technologies Related to Assistive Environments}, ~40.
\newblock ACM.

\bibitem[\protect\citeauthoryear{Reiss and Stricker}{2012b}]{reiss2012iswc}
Reiss, A., and Stricker, D.
\newblock 2012b.
\newblock Introducing a new benchmarked dataset for activity monitoring.
\newblock In {\em Wearable Computers (ISWC), 2012 16th International Symposium
  on},  108--109.
\newblock IEEE.

\bibitem[\protect\citeauthoryear{Rezende, Mohamed, and
  Wierstra}{2014}]{rezende2014arxiv}
Rezende, D.~J.; Mohamed, S.; and Wierstra, D.
\newblock 2014.
\newblock Stochastic backpropagation and approximate inference in deep
  generative models.
\newblock {\em arXiv preprint arXiv:1401.4082}.

\bibitem[\protect\citeauthoryear{Ribeiro, Saverese, and
  Figueiredo}{2017}]{ribeiro2017kdd}
Ribeiro, L.~F.; Saverese, P.~H.; and Figueiredo, D.~R.
\newblock 2017.
\newblock struc2vec: Learning node representations from structural identity.
\newblock In {\em Proceedings of the 23rd ACM SIGKDD International Conference
  on Knowledge Discovery and Data Mining},  385--394.
\newblock ACM.

\bibitem[\protect\citeauthoryear{Srivastava, Mansimov, and
  Salakhudinov}{2015}]{srivastava2015icml}
Srivastava, N.; Mansimov, E.; and Salakhudinov, R.
\newblock 2015.
\newblock Unsupervised learning of video representations using lstms.
\newblock In {\em International Conference on Machine Learning},  843--852.

\bibitem[\protect\citeauthoryear{Tang \bgroup et al\mbox.\egroup
  }{2015}]{tang2015www}
Tang, J.; Qu, M.; Wang, M.; Zhang, M.; Yan, J.; and Mei, Q.
\newblock 2015.
\newblock Line: Large-scale information network embedding.
\newblock In {\em Proceedings of the 24th International Conference on World
  Wide Web}.

\bibitem[\protect\citeauthoryear{Vincent \bgroup et al\mbox.\egroup
  }{2010}]{vincent2010jmlr}
Vincent, P.; Larochelle, H.; Lajoie, I.; Bengio, Y.; and Manzagol, P.-A.
\newblock 2010.
\newblock Stacked denoising autoencoders: Learning useful representations in a
  deep network with a local denoising criterion.
\newblock {\em Journal of Machine Learning Research} 11:3371--3408.

\bibitem[\protect\citeauthoryear{Wang and Gupta}{2015}]{wang2015iccv}
Wang, X., and Gupta, A.
\newblock 2015.
\newblock Unsupervised learning of visual representations using videos.
\newblock In {\em Proceedings of the IEEE International Conference on Computer
  Vision},  2794--2802.

\bibitem[\protect\citeauthoryear{Yang \bgroup et al\mbox.\egroup
  }{2017}]{yang2017icml}
Yang, B.; Fu, X.; Sidiropoulos, N.~D.; and Hong, M.
\newblock 2017.
\newblock Towards k-means-friendly spaces: Simultaneous deep learning and
  clustering.
\newblock In {\em Proceedings of the 34th international conference on Machine
  learning}.

\bibitem[\protect\citeauthoryear{Yeh, Kavantzas, and Keogh}{2017}]{yeh2017icdm}
Yeh, C.-C.~M.; Kavantzas, N.; and Keogh, E.
\newblock 2017.
\newblock Matrix profile vi: Meaningful multidimensional motif discovery.
\newblock In {\em 2017 IEEE 17th International Conference on Data Mining
  (ICDM)}.

\bibitem[\protect\citeauthoryear{Yeh}{2018}]{nnwebsite}
Yeh, C.-C.~M.
\newblock 2018.
\newblock Project website.
\newblock https://sites.google.com/view/neighbor-encoder/.

\bibitem[\protect\citeauthoryear{Zheng, Noroozi, and Yu}{2017}]{zheng2017joint}
Zheng, L.; Noroozi, V.; and Yu, P.~S.
\newblock 2017.
\newblock Joint deep modeling of users and items using reviews for
  recommendation.
\newblock In {\em Proceedings of the Tenth ACM International Conference on Web
  Search and Data Mining},  425--434.
\newblock ACM.

\end{thebibliography}
\bibliographystyle{aaai}
\end{document}